\documentclass{article}

\usepackage{microtype}
\usepackage{graphicx}
\usepackage{subfigure}
\usepackage{booktabs} 
\usepackage{amsmath}
\usepackage{graphicx} 
\usepackage{amsthm}
\usepackage{amsmath}
\usepackage{amssymb}
\usepackage{graphicx}
\usepackage{titling}
\usepackage{enumitem}
\usepackage{array}
\usepackage{multirow}
\usepackage{centernot}
\usepackage{float}
\usepackage{caption}
\usepackage{mathtools, amssymb, amsthm, bm}
\usepackage{amsmath}
\usepackage{titling}
\usepackage{etoolbox}
\usepackage{hyperref}
\usepackage{wrapfig}
\usepackage{caption}
\DeclareMathAlphabet\mathbfcal{OMS}{cmsy}{b}{n}

\DeclareMathOperator*{\argmin}{arg\,min}

\newtheorem{theorem}{Theorem}[section]

\newtheorem{lemma}[theorem]{Lemma}

\theoremstyle{definition}
\newtheorem{definition}{Definition}[section]

\usepackage{hyperref}

\usepackage[accepted]{icml2024}

\icmltitlerunning{Wasserstein Wormhole}

\begin{document}

\twocolumn[
\icmltitle{Wasserstein Wormhole: Scalable Optimal Transport Distance with Transformers}



\icmlsetsymbol{equal}{*}

\begin{icmlauthorlist}
\icmlauthor{Doron Haviv}{MSKCC,CBM}
\icmlauthor{Russell Zhang Kunes}{MSKCC,Columbia}
\icmlauthor{Thomas Dougherty}{MSKCC,CBM}
\icmlauthor{Cassandra Burdziak}{MSKCC}
\icmlauthor{Tal Nawy}{MSKCC}
\icmlauthor{Anna Gilbert}{Yale}
\icmlauthor{Dana Pe'er}{MSKCC,HHMI}
\end{icmlauthorlist}

\icmlaffiliation{MSKCC}{Computational and Systems Biology Program, Sloan Kettering Institute, Memorial Sloan Kettering Cancer Center}

\icmlaffiliation{CBM}{Tri-Institutional Training Program in Computational Biology and Medicine, Weill Cornell Medicine}
\icmlaffiliation{HHMI}{Howard Hughes Medical Institute}
\icmlaffiliation{Yale}{Department of Mathematics and Statistics, Yale University}
\icmlaffiliation{Columbia}{Department of Statistics, Columbia University}
\icmlcorrespondingauthor{Dana Peer}{peerd@mskcc.com}

\icmlkeywords{Machine Learning, ICML}

\vskip 0.3in
]



\printAffiliationsAndNotice{} 

\begin{abstract}

Optimal transport (OT) and the related Wasserstein metric ($W$) are powerful and ubiquitous tools for comparing distributions. However, computing pairwise Wasserstein distances rapidly becomes intractable as cohort size grows. An attractive alternative would be to find an embedding space in which pairwise Euclidean distances map to OT distances, akin to standard multidimensional scaling (MDS). We present Wasserstein Wormhole, a transformer-based autoencoder that embeds empirical distributions into a latent space wherein Euclidean distances approximate OT distances. Extending MDS theory, we show that our objective function implies a bound on the error incurred when embedding non-Euclidean distances. Empirically, distances between Wormhole embeddings closely match Wasserstein distances, enabling linear time computation of OT distances. Along with an encoder that maps distributions to embeddings, Wasserstein Wormhole includes a decoder that maps embeddings back to distributions, allowing for operations in the embedding space to generalize to OT spaces, such as Wasserstein barycenter estimation and OT interpolation. By lending scalability and interpretability to OT approaches, Wasserstein Wormhole unlocks new avenues for data analysis in the fields of computational geometry and single-cell biology. Software is available at \url{http://wassersteinwormhole.readthedocs.io/en/latest/}.

\end{abstract}

\section{Introduction}

Many problems in statistical learning boil down to computing and comparing probability distributions that fit empirical data, and different approaches vary in their attention to distribution shape, spread, and degree of parameterization. Optimal transport (OT) methods attempt to find an ideal pairing between individual data points from two distributions \cite{villani2009optimal} by representing distances as the minimum cost of mapping between (weighted) point clouds. While original OT solvers, which rely on linear programming, scale poorly \cite{peyre2019computational}, newer approaches can reduce the computational complexity and generalize OT to new domains such as computational biology \cite{nitzan2019gene, schiebinger2019optimal, bunne2023learning}, numerical geometry \cite{su2015optimal}, chemistry \cite{wu2023improving}, and image processing \cite{feydy2017optimal}. 



We are specifically interested in massive cohorts of empirical distributions, each with up to thousands of points. The computational complexity of finding the Wasserstein distance between two point clouds with Sinkhorn's algorithm \cite{cuturi2013sinkhorn}, the current favored \cite{flamary2021pot} OT solver, is $O(n^{2} \cdot I)$, where $n$ is the size of each point cloud and $I$ is the number of Sinkhorn iterations until convergence. As we must repeat Sinkhorn's algorithm for all pairs in a cohort, requiring $O(N^2 \cdot n^2 \cdot I)$ time to analyze $N$ distributions, current OT algorithms are unable to analyze large cohorts despite improvements in scalability.



 
Here, we present Wasserstein Wormhole, an algorithm that represents each point cloud as a single embedded point, such that the Euclidean distance in the embedding space matches the OT distance between point clouds. In Wormhole space, we compute Euclidean distance in $O(d)$ time for an embedding space with dimension $d$, which acts as an approximate OT distance and enables Wasserstein-based analysis without expensive Sinkhorn iterations. 

Point clouds contain a variable number of unordered samples, hence neural architectures that require a fixed-sized or ordered input are not suited to encode them. On the contrary, transformers \cite{vaswani2017attention} are uniquely poised to do so, due to their permutation equivariance, indifference to input length, and ability to model relationships between samples in each point cloud via attention. 

\begin{figure*}[h]
  \centering
  \includegraphics[width=0.90\textwidth]{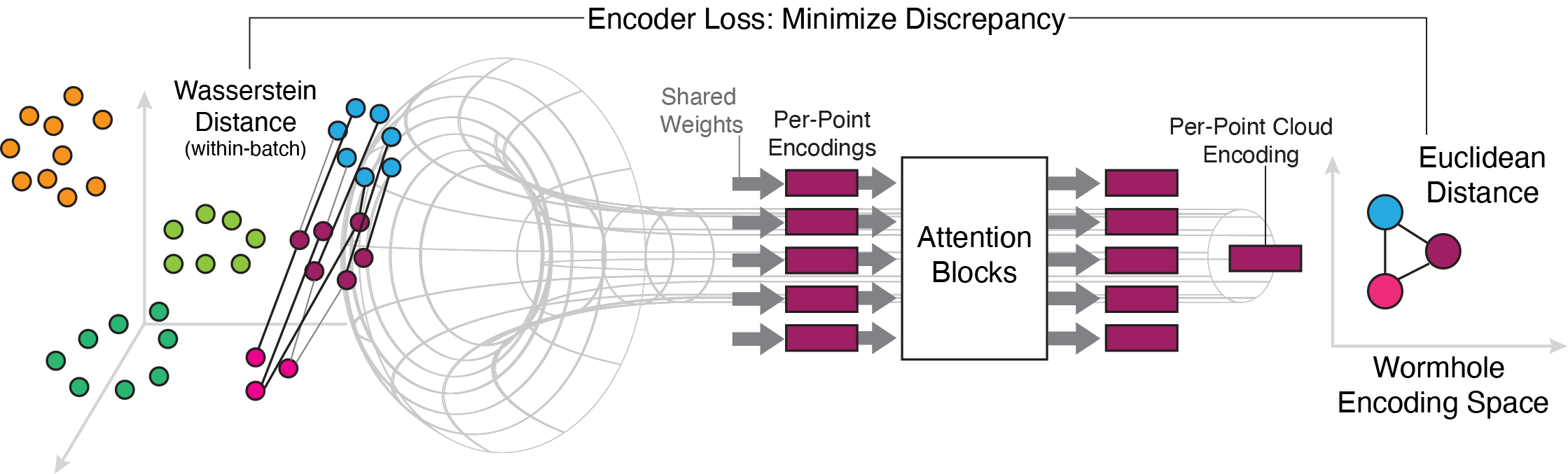}
  \caption{\textbf{Schematic of Wasserstein Wormhole}. Empirical distributions (point clouds) are passed through a transformer to produce per point-cloud vector embeddings such that the Euclidean distance between embeddings match the pairwise Wasserstein distance between point clouds. Since computation of OT distances is laborious, Wormhole is optimized by mini-batches to minimize the discrepancy between the embedding pairwise distances and the pairwise Wasserstein distances of the batch point clouds. The Wormhole decoder (not shown) is a second transformer trained to reproduce the input point clouds from the embedding by minimizing the OT distance between input and output.}
  \label{mainfig:cocept}
\vspace{-1.5em}
\end{figure*}

In addition to an encoder that embeds point clouds, Wormhole trains a decoder that reconstructs point clouds from their embedding. The decoder facilitates interpretability with its ability to explicitly compute Wasserstein barycenters \cite{cuturi2014fast} or interpolate between point clouds \cite{chewi2021fast}.

The problem solved by Wormhole is analogous to multidimensional scaling \cite{torgerson1952multidimensional}. Both methods find an embedding that preserves distances between samples. While several distance metrics are provably capable of being represented in this manner, it has been shown that a host of other "non-Euclidean" metrics, including Wasserstein, cannot be exactly equated with Euclidean distance. Counter-intuitively, classical MDS on non-Euclidean distance matrices becomes more inaccurate as embedding dimensions increase \cite{sonthalia2021can}.

Inspired by \cite{sonthalia2021can}, we extend MDS theory to derive upper and lower bounds for the error incurred by embedding non-Euclidean distances, such as Wasserstein. We further describe a projected gradient descent (PGD) \cite{boyd2004convex} algorithm for MDS which is guaranteed to converge to the global optimum for any distance matrix. On small examples where it is feasible to compute bounds and apply PGD, we find that Wormhole approaches the optimal solution and falls within our derived bounds.

\begin{table*}
\begin{center}
\scalebox{0.70}{
\begin{tabular}{|c|c|c|c|c|c|c|c|c|c|c|c|}
\hline
& \multicolumn{3}{c|}{Dataset Parameters}&
\multicolumn{4}{c|}{OT Correlation}&
\multicolumn{4}{c|}{Label Accuracy}\\
\hline
 Name       & Cohort Size & Cloud Size & Dimension & Wormhole & DiffusionEMD & DWE & Fréchet & Wormhole & DiffusionEMD & DWE & Fréchet \\[0.5ex] 
\hline
 MNIST      & 70,000    & 105 & 2  & \textbf{0.98} & 0.845& 0.92  & 0.86     & \textbf{0.98} & 0.84 & 0.97  & 0.49 \\
 Fashion-MNIST  & 70,000    & 356 & 2  & \textbf{0.99} & 0.97  & 0.98 & 0.99 & \textbf{0.87} & 0.77 & \textbf{0.87} & 0.73 \\
 ModelNet40    & 12,311    & 2048 & 3  & \textbf{0.99} & 0.85  & 0.97 & 0.95 & \textbf{0.86} & 0.69 & 0.78 & 0.61 \\
 ShapeNet     & 15,011    & 2048 & 3  & \textbf{0.99} & 0.81  & 0.97  & 0.97 & \textbf{0.98} & 0.94 & 0.97 & 0.92 \\
 \hline
 \hline
 Rotated ShapeNet (GW) & 15,011    & 512 & 3  & \textbf{0.98} & NA   & NA  & 0.68     & \textbf{0.82} & NA & NA & 0.16 \\
 \hline
 \hline
 MERFISH Cell Niches  & 256,058   & 11  & 254 & \textbf{0.97} & OOM  & OOM  & OOM      & \textbf{0.98} & OOM  & OOM & OOM \\
 SeqFISH Cell Niches  & 57,407   & 29 & 351 & \textbf{0.98} & OOM  & OOM  & OOM      & \textbf{0.97} & OOM  & OOM & OOM \\
 scRNA-seq Atlas  & 2,185   & 69  & 2500 & \textbf{0.98} & OOM  & OOM  & OOM      & \textbf{0.96} & OOM  & OOM & OOM \\
\hline
\end{tabular}}
\end{center}
\caption{Summary of each dataset used in this study and results from every benchmarked algorithm. Cloud size denotes the median number of points in each point cloud for every dataset. Pearson correlations are computed between true and estimated Wasserstein distances according to pairwise matrices of $10$ instances of $128$ randomly chosen point clouds from each dataset (OT Correlation). The mean squared error (MSE) follows the same trend with Wormhole outperforming competing methods (Table \ref{table:MSE}). Object classification accuracy is based on the embeddings of each algorithm. After computing encodings, a classifier was trained on the training set embeddings and we denote the classifier's accuracy on the test set (Label Accuracy). Rotated ShapeNet is a version of the standard ShapeNet dataset where each sample is randomly rotated in 3D, subsampled by a factor of $4$, and embedded via Gromov-Wasserstein (GW). We divide the results into Wasserstein on 2D and 3D point-clouds, Gromov-Wasserstein on rotated data with arbitrary orientation and high-dimensional datasets.
Full implementation details are described in the supplement (Section \ref{section:imp_details}). NA, Not Applicable. OOM, Out-Of-Memory.}
\label{table:summary}
\vspace{-1em}
\end{table*}



\section{Prior Work}
There have been many prior works on learning from point cloud data \cite{guo2020deep}, often involving OT-based solutions. PointNet \cite{garcia2016pointnet, qi2017pointnet} introduced the idea of permutation-invariant, fully-connected neural networks to classify and segment 3D point-clouds, which was later extended into transformer-based networks \cite{lee2019set}. In the context of OT, several works have noted the expressiveness of Wasserstein space compared to euclidean space \cite{frogner2019learning, katageri2024metric, liu2022wasserstein}, and have developed methods to embed point-clouds into Wasserstein space, from which relevant information can be more reliably extracted. These works are in essence the inverse of Wormhole, as their target space is Wasserstein, in contrast to Wormhole's Euclidean target space.

\begin{figure}[h]
  \centering
    \includegraphics[width=0.42\textwidth]{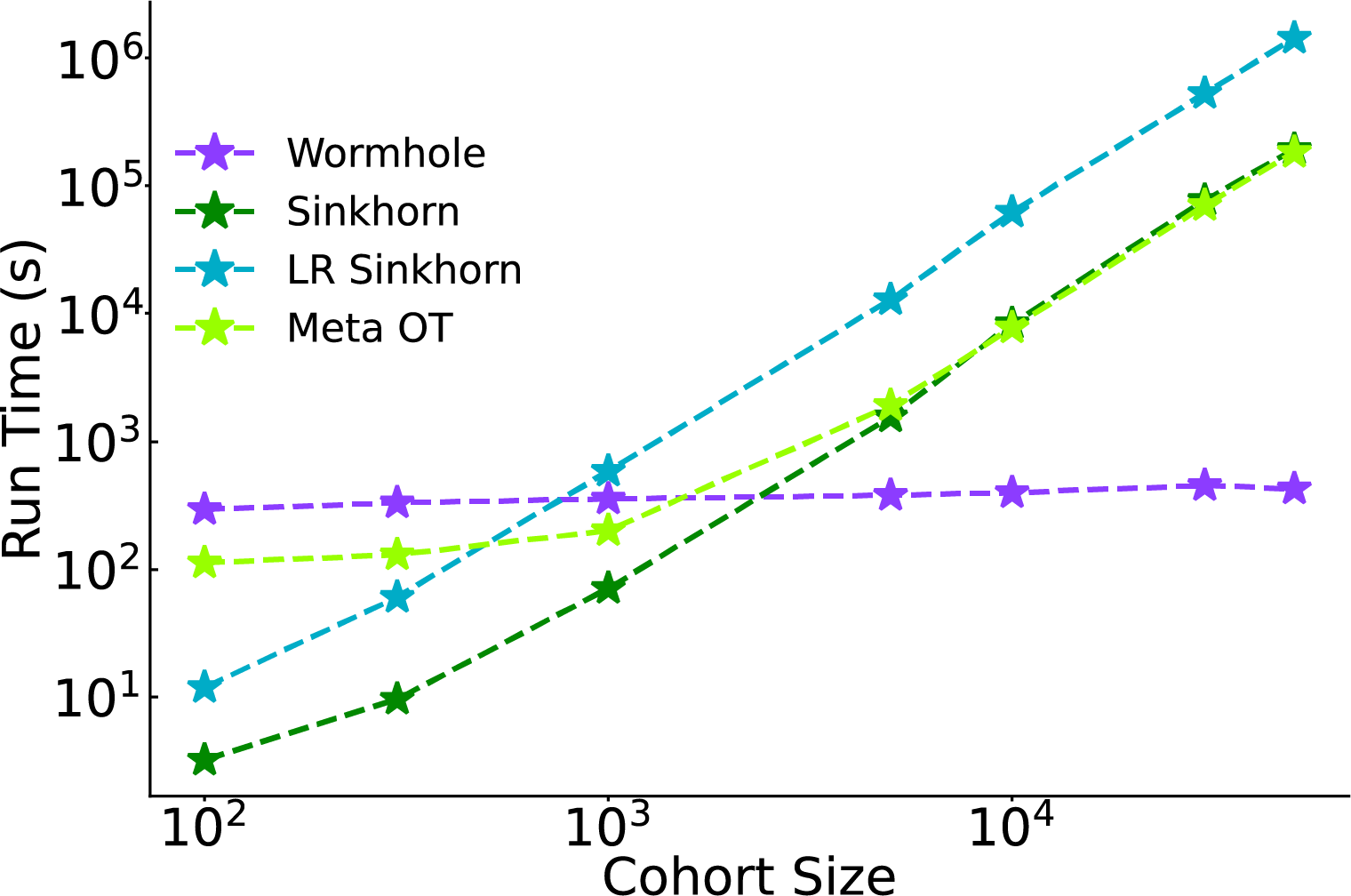}
    \caption{\textbf{Benchmarking run-time of Wormhole against other OT acceleration algorithms}. Sampling cohorts of different sizes from the MNIST dataset, we measured the time required for current acceleration algorithms (with GPU implementation in JAX-OTT) and Wormhole to compute or approximate the pairwise Wasserstein matrix. Other methods are more appropriate at tiny cohorts, as they do not require training a parametric model. However, even in cohorts with relatively few samples, Wormhole is superior and no other method can scale to complete datasets, requiring weeks of compute time, even on a fully-utilized 80GB GPU.}
    \label{mainfig:time_benchmark}
\end{figure}
\vspace{-0.5em}

Wasserstein Wormhole belongs to a class of algorithms that exploit structures within distributions or relationships between cohorts of them to accelerate or approximate OT. Low-Rank (LR) Sinkhorn \cite{scetbon2021low} factorizes the pairing matrix into a low-rank composition, assuming the transport plan between two distribution is structured, to enable faster computation of Sinkhorn iterations. Convolutional and Geodesic Sinkhorn \cite{bachmann2022wasserstein, solomon2015convolutional} construct solvers which leverage the underlying geometry of the point-cloud space to simplify OT optimization. MetaOT \cite{amos2022meta} recognizes that convergence of Sinkhorn depends greatly on quality of initialization and fits a neural network which learns an initial Sinkhorn potential for a given pair of input distributions. Due to its fully-connected architecture, MetaOT is constrained to distributions which share supports, and cannot be applied to arbitrary point-clouds. Although these methods can speed up the computation of an individual OT problem, none alleviate the need for many expensive pairwise computations, and are far slower than Wormhole on large cohorts (Figure \ref{mainfig:time_benchmark}). 

Previous works have also employed Wasserstein preserving euclidean embeddings in the pursuit of scaling OT to large cohorts \cite{tong2021diffusion}, some using neural network models \cite{courty2017learning, sehanobish2020permutation}. DWE \cite{courty2017learning} trains a Siamese neural network to match distances, and DiffusionEMD \cite{tong2021diffusion} extends the wavelet-based approximation of EMD \cite{shirdhonkar2008approximate} onto graph domains. Unlike our Wormhole approach, which applies to point clouds in any dimension, DWE is constrained to small 2D images or 3D grids, while scalability of DiffusionEMD is limited to distributions that share supports, such as grayscale images or voxelized spaces. We benchmark Wormhole against DiffusionEMD \& DWE, and show that our method is superior at matching OT distances and deriving insights from embeddings on low-dimensional point clouds. On general datasets, Wormhole is the only embedding method which can scale to high-dimensions.

\section{Preliminaries}\label{section:Preliminaries}

\subsection{Optimal Transport}

We follow the formulation of optimal transport presented in \cite{peyre2019computational}. Given two discrete distributions $P_{x}=\sum_{i=1}^{n}\mu_{i}\delta_{x_{i}}$ and $P_{y}=\sum_{i=1}^{m}\nu_{i}\delta_{y_{i}}$ ($\delta$ is the Dirac function), their entropically regularized OT distance is:

\begin{equation}\label{eq:w2}
  W_{\varepsilon}(P_{x}, P_{y}) = \min_{P\in U(\nu,\mu)}\sum_{i,j} c(x_{i}, y_{j})P_{i,j}\ - \varepsilon H(P)
\end{equation} 

Where $U(\nu,\mu) := \{ P \in R_{+}^{n,m}: \sum_j P_{i,j} = \nu_{i} ,\sum_i P_{i,j} = \mu_{j}\}$ is the set of all valid pairing matrices, $c(x_{i}, y_{j})$ is the ground cost function between points in $P_{x}$ and $P_{y}$, usually chosen to be the Euclidean distance, and $H(P) = -\sum_{i,j}P_{i,j}(\log P_{i,j}-1)$ is the entropic term. While the unregularized problem incurs a heavy computational burden for distributions with large supports or in high dimensions, $W_{\varepsilon}$ can be computed via a series of GPU-accelerable vector-matrix products known as Sinkhorn iterations \cite{cuturi2013sinkhorn}.

Due to the regularization term, the $W_{\varepsilon}$ distance between a distribution and itself is not 0, which can introduce biases when used to optimize generative models. To allow for effective optimization while retaining the computational advantages of entropic regularization, a common alternative is the Sinkhorn Divergence \cite{genevay2018learning}, which subtracts the self-transport costs from $W_{\varepsilon}$:

\begin{equation}\label{eq:s2}
  S_{\varepsilon}(P_{x}, P_{y}) = W_{\varepsilon}(P_{x}, P_{y}) - \frac{W_{\varepsilon}(P_{x}, P_{x}) + W_{\varepsilon}(P_{y}, P_{y})}{2}
\end{equation} 

\subsection{Multi-dimensional Scaling}

Given the pairwise Wasserstein distance matrix between a cohort of distributions $D_{i,j} = W_{\varepsilon}(X_{i}, X_{j})$, we seek an embedding per distribution, such that pairwise Euclidean distance between embeddings match the OT distance $\lVert \alpha_{i} -\alpha_{j} \rVert_{2}^{2}\approx W_{\varepsilon}(X_{i}, X_{j}))$. In essence, this describes the MDS problem with the critical difference that for large cohorts, it is computationally infeasible to compute the OT distance between all pairs of distributions. Still, MDS can provide key theoretical insights for our setting. 

\begin{definition}[Euclidean distance matrix (EDM)]\label{EDM:def}
An $N\times N$ distance matrix $D$ is Euclidean if there exists a set of points $X = \{x_{1}, x_{2},\dots,x_{N}\}$ in some dimension $x_{i}\in R^{d}$ such that:

\begin{equation}
  \forall{i,j}: \; D_{i,j} = \| x_{i} - x_{j} \|_{2}^2 
\end{equation}

From \cite{GOWER198581}, a matrix $D$ is a Euclidean Distance Matrix if and only if all of its values are non-negative, it is hollow (diagonal elements are 0) and the criterion matrix $F$ is positive-semidefinite (PSD):

\begin{equation}\label{eq:jdj_psd}
  F = -JDJ
\end{equation}

Where $J = I_{N} -\frac{\mathbf{1}_{N}\mathbf{1}_{N}^{\top}}{N}$ is the centering matrix and $1_{N}$ is a vector of length $N$ filled with ones. 

\end{definition}

Wasserstein distance is not Euclidean (see section 8.3 in \cite{peyre2019computational} and \cite{naor2007planar}), and there are many examples of cohorts of distributions (Figures \ref{suppfig:TheoryPGDFigure},\ref{suppfig:mnist_spectrum}) wherein pairwise OT distance matrices do not fit the criteria in equation \ref{eq:jdj_psd}. Unlike the Euclidean scenario where classical MDS (cMDS) improves as embedding size grows, on non-Euclidean distance matrices the error behaves counter-intuitively and worsens as dimension increases \cite{sonthalia2021can}. This motivates us to seek a description of the theory underpinning the embeddings of non-Euclidean distances.

\section{Wasserstein Wormhole}\label{section:algorithm}

The complexity of modern OT solvers is quadratic with respect to the number of points in a point cloud; thus, it is only straightforward to compute pairwise Wasserstein distance for small cohorts of empirical distributions. Indeed, naive approaches will not even scale to standard datasets, such as the 12,311-point-cloud ModelNet40 benchmark  \cite{wu20153d}.

This problem calls for a stochastic approach. Instead of directly assigning each point cloud with an embedding, we propose learning a parametric encoding function. Formally, the dataset in our setting is a cohort of point clouds $\Omega = \{X_{1}, X_{2}, X_{3}, \dots X_{N}\}$, where each point cloud $X_{i}$ contains $n_{i}$ points in $d$ dimensions. Our goal is to find a mapping function $T(X)$ that encodes point clouds into embeddings $\alpha$, such that Euclidean distances between encodings approximate the OT distances been point clouds. By computing the pairwise Wasserstein distance matrix for small mini-batches, we can train $T(X)$ so that encoding distances directly match Wasserstein.


Since $T(X)$ is applied to point clouds, it should contain a few key attributes:

\setlist{nolistsep}
\begin{itemize}[noitemsep]
 \item Wasserstein distance is invariant with respect to the order of samples within point clouds ($W(X, PY) = W(X,Y)$, $P$ is a permutation matrix). The encoding must therefore also be permutation invariant: $T(X) = T(PX)$. 
 \item Point clouds contain varying numbers of samples. $T(X)$ must be able to generalize to arbitrary cloud sizes while modeling relationships between the points within them.
\end{itemize}

Transformers—specifically, stacked layers of multi-head attention—fit both of these criteria. Although they are commonly associated with sequential data such as words in a sentence, order information is provided as input to the model. Without the positional embeddings, attention mechanisms (and therefore transformers) are permutation equivariant (proof in supplementary section \ref{section:permutation}). By performing an invariant function such as averaging on the final set of embeddings, this produces a mapping $T(X)$ that is also invariant by design. 

\begin{algorithm}[h]
\caption{\textbf{Wasserstein Wormhole} At each training step, a mini-batch of point clouds is randomly sampled from the cohort. Wormhole is optimized so that pairwise distance between mini-batch embeddings match their OT distance and output decodings reconstruct input point clouds.}\label{alg:Wormhole}
\begin{algorithmic}
  \STATE {\bfseries Input:} point clouds $\Omega=\{X_{i}\}^{N}_{i=1}$, initialized encoder $T$ and decoder $G$ networks, batch size $B$, learning rate $\varepsilon_{t}$ 
  \REPEAT
  \STATE Sample $B$ point clouds $\{x_{i}\}_{i=1}^{B}$ from $\Omega$ 
  \STATE Calculate encodings $\alpha_{i} = T(x_{i})$
  \STATE Compute pairwise OT in batch $D_{i,j}=S_\varepsilon(x_{i},x_{j})$
  \STATE Evaluate stress $\mathcal{L}_{enc}=\sum_{i,j}(\lVert \alpha_{i} -\alpha_{j} \rVert_{2}^{2} - D_{i,j})^2$
  \STATE Produce decodings $\hat{x}_{i} = G(\alpha_{i})$
  \STATE Evaluate decoding error $\mathcal{L}_{dec}=\sum_{i}S_\varepsilon(x_{i},\hat{x}_{i})$
  \STATE Update encoder $T \leftarrow T - \varepsilon_{t}\nabla(\mathcal{L}_{enc}+\mathcal{L}_{dec})$
  \STATE Update decoder $G \leftarrow G - \varepsilon_{t}\nabla(\mathcal{L}_{dec})$
  \UNTIL{Convergence}
\end{algorithmic}
\end{algorithm}

By training the encoder network $T(X)$ (Figure \ref{mainfig:cocept}, Algorithm \ref{alg:Wormhole}), we can produce encodings which allow for OT-based analysis of point clouds in linear time, enabling methods such as clustering and visualization \cite{bachmann2022wasserstein} on the entire cohort. However, the encoder alone does not provide a clear interpretation of these analyses. 

 \begin{figure}[h]
  \centering
  \includegraphics[width=0.48\textwidth]{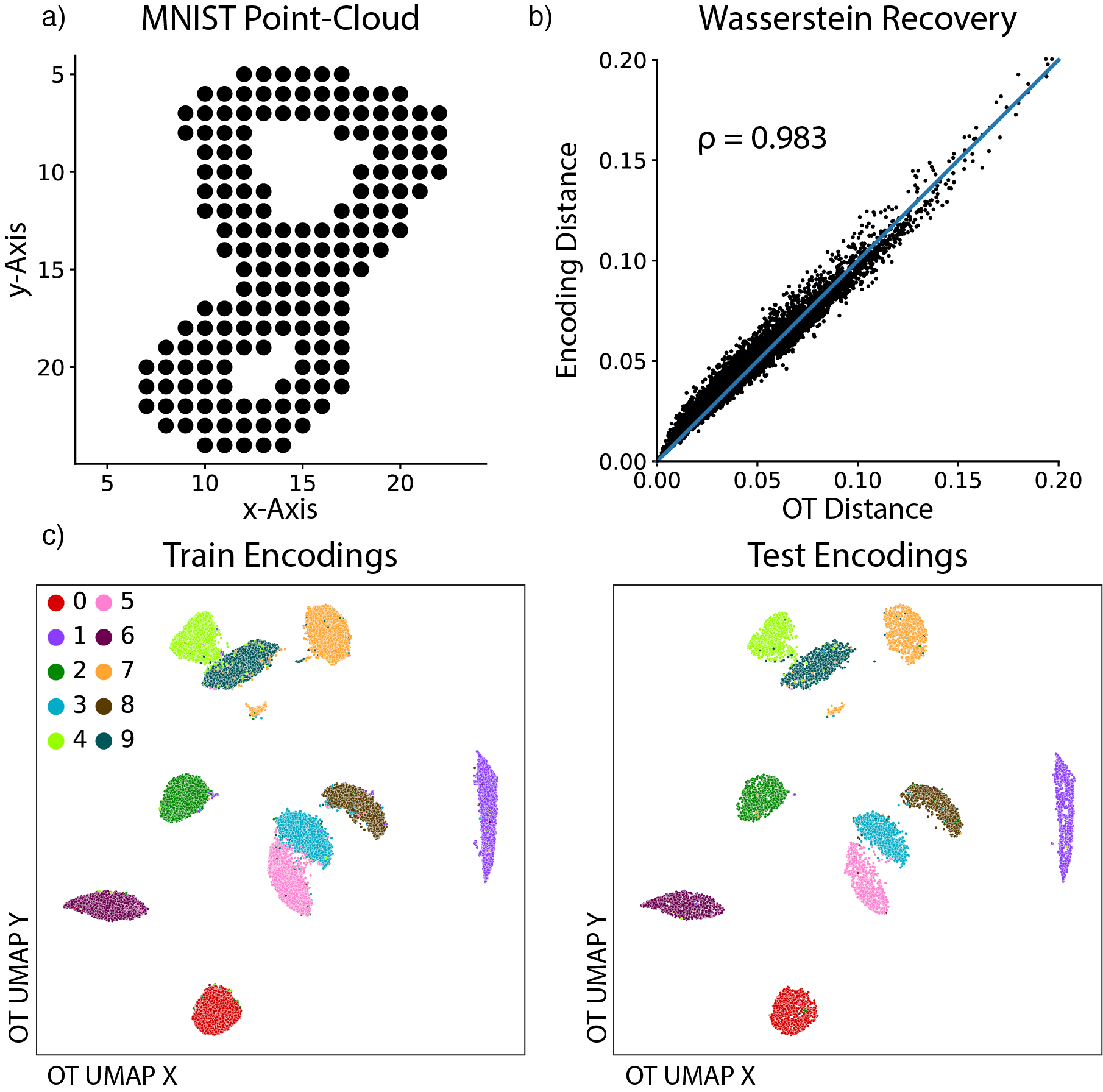}
  \caption{\textbf{Wormhole on MNIST point clouds. a}. Example of a point cloud produced by thresholding MNIST images. \textbf{b.} Retrieval of the Wasserstein distance in Wormhole embedding space. From a random sample of $128$ point clouds from the test set, we denote the correlation between the true pairwise OT distance and the Euclidean distance between their embeddings. The line $y=x$ is drawn in blue for reference. \textbf{c.} 2D UMAP visualization of Wormhole embeddings of the training and test set point clouds, which recapitulates ground-truth image digit without utilizing label information during training.}
  \label{mainfig:MNIST}
\end{figure}

To imbue the embeddings with interpretability, we include a decoder network $G(\alpha)$ which takes encodings as input and is simultaneously trained to reproduce the input point clouds by minimizing the Wasserstein distance loss. Unlike standard autoencoders, which only train the full network to match outputs to inputs, Wormhole imposes a loss on both the encodings and decodings.

\section{Embedding Non-Euclidean Distance Matrices}\label{section:theory}

Given a distance matrix $D$ sized $N\times N$, we are interested in how well it can be approximated with a series of embeddings $\{\alpha_{i}\}_{i=1}^{N}$ in $d$ dimensions $\alpha_{i}\in R^{d}$. This corresponds to the optimization problem of minimizing the cumulative \textit{stress}:

\begin{equation}\label{eq:MDS_goal}
\mathcal{L} =  \min_{\{\alpha_{1},\alpha_{2},\dots,\alpha_{N}\}}\sum_{i,j}(\lVert \alpha_{i} -\alpha_{j} \rVert_{2}^{2} - D_{i,j})^2 \quad 
\end{equation} 

In the case where $D$ is Euclidean, by definition there exists a set of embeddings in dimension $d = N$ where the distance matrix is perfectly reconstructed and $\mathcal{L}=0$. Wormhole, however, is trained to minimize the \textit{stress} relative to the Wasserstein distance matrix, which is generally \textit{not} Euclidean. It is currently unclear to what extent $D$ can be represented with embeddings when it is not Euclidean \cite{sonthalia2021can}.


As formulated in equation \ref{eq:MDS_goal}, this problem is non-convex and therefore difficult to study. However, when we instead reparameterize the objective to be a function over the distance matrix produced from the embedding $\hat{D}_{i,j} = \lVert \alpha_{i} -\alpha_{j} \rVert_{2}^{2}$ and add relevant constraints for $\hat{D}$ to be Euclidean, the problem is rendered convex :

\begin{equation}\label{eq:convex_MDS}
\begin{gathered}
  \mathcal{L} = \min_{\hat{D}}\sum_{i,j}(\hat{D}_{i,j} - D_{i,j})^2 \\
  \text{s.t. $C_{1}$: } -J\hat{D}J \text{ is PSD, $C_{2}$: } \hat{D} \text{ is hollow and $C_{3}$: } \hat{D}_{i,j}\geq0.
\end{gathered}
\end{equation}

Since all constraints are convex sets, we could construct a projected gradient descent approach (Algorithm \ref{alg:Projected}) that is guaranteed to converge to the global minimum $\mathbfcal{L^{*}}$. Furthermore, we derive novel theoretical lower and upper bounds for the global minimum $\mathbfcal{L^{*}}$.

\begin{theorem}[Lower Bound]
\label{theorem:lower_bound}
For a given distance matrix $D$ and the eigendecomposition $\{\lambda_{i}, v_{i}\}_{i=1}^{N}$ of its criterion matrix $F=-JDJ$, the optimal \textit{stress} is greater or equal to:
\begin{equation}\label{eq:lower_bound}
\mathbfcal{L^{*}} \geq \mathbf{L} = \sum_{i: \lambda_{i}<0} \lambda^{2}_{i}
\end{equation} 
\end{theorem}

This bound also appears in \cite{sonthalia2021can}, which analyzed pathologies of cMDS on non-Euclidean distance matrices.

\begin{theorem}[Upper Bound]
\label{theorem:upper_bound}
The optimal \textit{stress} is less than or equal to:
\begin{equation}\label{eq:upper_bound}
\mathbfcal{L^{*}} \leq \mathbf{U} = \sum_{i,j}(\Delta g_{i}+\Delta g_{j})^2 + \sum_{i: \lambda_{i}<0}\lambda_{i}^2
\end{equation} 
Where:
\begin{equation}
\Delta g_{i} = \frac{1}{2}\sum_{j: \lambda_{j}<0}\lambda_{j}\cdot v_{i,j}^2
\end{equation}

And $v_{i,j}$ is the $j$-th element of the $i$-th eigenvector of $F$.

\end{theorem}

The prominence of the negative spectrum of the criterion matrix can be thought of as a quantitative measure of how close $D$ is to being Euclidean. As we expect, the more minute the negative eigenvalues of $F$ (as in Figure \ref{suppfig:mnist_spectrum}), the smaller both the lower and upper bounds become, and we can anticipate that the embeddings will increase in quality.

The derivation of the lower bound comes from removing the constraints $C_{1}$ and $C_{2}$ on hollowness and non-negativity of $\hat{D}$ and computing the projection operator onto $C_{1}$ which is the space of matrices where $-JDJ$ is positive semi-definite (subsection \ref{subsection:lower_bound}). Although it is similar to the standard projection onto PSD and incurs the same truncation error, it has additional terms due to the double-centering transform. 

\begin{algorithm}
\caption{Theoretically Optimal Embedding of non-Euclidean Distance Matrices with Projected Gradient Descent (\textbf{PGD})}
  \label{alg:Projected}
\begin{algorithmic}
  \STATE {\bfseries Input:} Distance Matrix $D$, Initial Solution $D'_{0}$, Learning Rate $\varepsilon_{t}$ 
  \REPEAT
  \STATE $D^{*}_{t+1} = D'_{t} - \varepsilon_{t}\nabla( \|D'_{t} - D\|_F^2)$ \COMMENT{GD Step}
  \STATE $D'_{t+1} = \text{proj}_{C_{1} \cup C_{2} \cup C_{3}}(D^{*}_{t+1})$ \COMMENT{Dykstra}
  \UNTIL{Convergence}
  \STATE{\textbf{RETURN}} $X = \text{cMDS}(D')$ \COMMENT{Embed $D'$ with MDS}
\end{algorithmic}
\end{algorithm}

This new projection operator, along with the simple projections onto $C_{2}$ (hollow matrices, by setting the diagonal to $0$) and $C_{3}$ (non-negative matrices, by thresholding), allows us to formulate a PGD \cite{boyd2004convex} scheme (Algorithm \ref{alg:Projected}) to embed non-Euclidean distance matrices. This algorithm projects onto $C_{1}\cup C_{2}\cup C_{3}$ using Dykstra iterations \cite{boyle1986method} and is guaranteed to find the optimum for equation \ref{eq:convex_MDS}, unlike other MDS approaches such as cMDS or SMACOF \cite{de2009multidimensional}. The PGD algorithm requires knowing the full pairwise distance matrix and is therefore infeasible in our setting of massive cohorts. However, for small cohorts, the theoretical assurances of PGD provide a useful benchmark for Wormhole.

Our analysis indicates that in real-world datasets, the Wormhole approximation is highly accurate and commensurate with theoretical values. Although this assessment is based on a small subsample of the full dataset, we assume that if random subsets of the larger dataset can be well embedded, the embedding of the entire dataset is reliable (Table \ref{table:bounds}). Full derivation of the theoretical bounds appear in Supplementary Sections \ref{section:theory_supp} \& \ref{subsection:ProjectedGradientDescent}.

\begin{table}[h]
\scalebox{0.85}{
\begin{tabular}{l c c c c c}
 \hline
 Dataset  & Lower & Upper & PGD & Wormhole & cMDS \\ [0.5ex] 
 \hline\hline
 Simplex ($35$) & $0.765$& $6.369$& $1.117$&  $1.420$ & $7.134$ \\
 Gaussian ($128$)  & $0.129$& $2.552$& $0.168$&  $0.401$ & $2.681$  \\
 MNIST ($256$)   & $0.042$& $0.616$& $0.058$&  $0.100$ & $0.657$  \\
 \hline
\end{tabular}}
\caption{Comparison between embedding algorithms and theoretical bounds on small datasets where full Wasserstein distance matrices are realizable. Values are the \textit{stress} between the embedding and true distances and parentheses denote the size of each dataset. Classical MDS produces far from optimal errors, falling outside our bounds. Wormhole training is stochastic, yet it still converges between the error bounds and approaches the error obtained by the theoretically optimal PGD. Both PGD and Wormhole embeddings were well correlated ($\rho=0.99$) with OT distances.}\label{table:bounds}
\end{table}

\section{Experiments}

We demonstrate Wasserstein Wormhole on datasets representing multiple contexts where OT is frequently applied. We applied Wormhole to 2D point clouds derived from the MNIST and Fashion-MNIST datasets, and to 3D CAD models from ModelNet40 and ShapeNet \cite{chang2015shapenet}. As a high-dimensional applications, we used Wormhole to embed multicellular niches \cite{haviv2023covariance} from two high-resolution spatial transcriptomics datasets \cite{zhang2021spatially, lohoff2022integration}, along with groups of cells ('MetaCells' \cite{persad2023seacells}) from a scRNA-seq atlas of patient response to COVID \cite{stephenson2021single}.

For each dataset, we quantitatively benchmarked Wormhole by measuring how well the learned embeddings matched the true OT distance and ground-truth labels. Since the full Wasserstein distance matrix is unattainable, for each dataset we randomly sampled $10$ instances of $n=128$ point clouds and compared the $n\times n$ OT distance matrix to the embedding Euclidean distance. All the datasets are labeled, allowing us to measure the classification accuracy based on held-out test-sets for each.

We benchmarked our Wormhole against DiffusionEMD and Deep Wasserstein Embedding (DWE), both sharing Wormhole's goal of finding OT preserving embeddings. We also compare against a constructed baseline we dub \textit{Fréchet} \cite{dowson1982frechet} in which we use the closed-form $W_{2}$ distance for Gaussian distributions. For each point cloud, we calculated the mean ($\mu$) and covariance matrix ($\Sigma$) of its samples and use the following formula to approximate the OT distance between each point cloud pair:

\begin{gather}\label{eq:frechet}
W_{2}^{2}\approx \| \mu_{i} - \mu_{j} \|^2 + \text{Tr}\left(\Sigma_{i} + \Sigma_{j} - 2\sqrt{\Sigma_{i}\Sigma_{j}}\right)
\end{gather}

The following sections present vignettes of Wormhole from several dataset, and results are summarized in Table \ref{table:summary}. Wormhole outperforms all competing approaches and uniquely scales to massive \& high-dimensional datasets such as the the 256,058-sample, 254-dimensional \textit{MERFISH Cell Niches} cohort. We specifically highlight the improvement of Wormhole over DWE, which similarly learns a neural network parametric embedding. The transformer backbone of Wormhole is superior to DWE's CNN in low-dimensional datasets, while also proving to be essential in scaling Wormhole beyond 3D data.

\begin{figure}[h]
  \centering
  \includegraphics[width=0.35\textwidth]{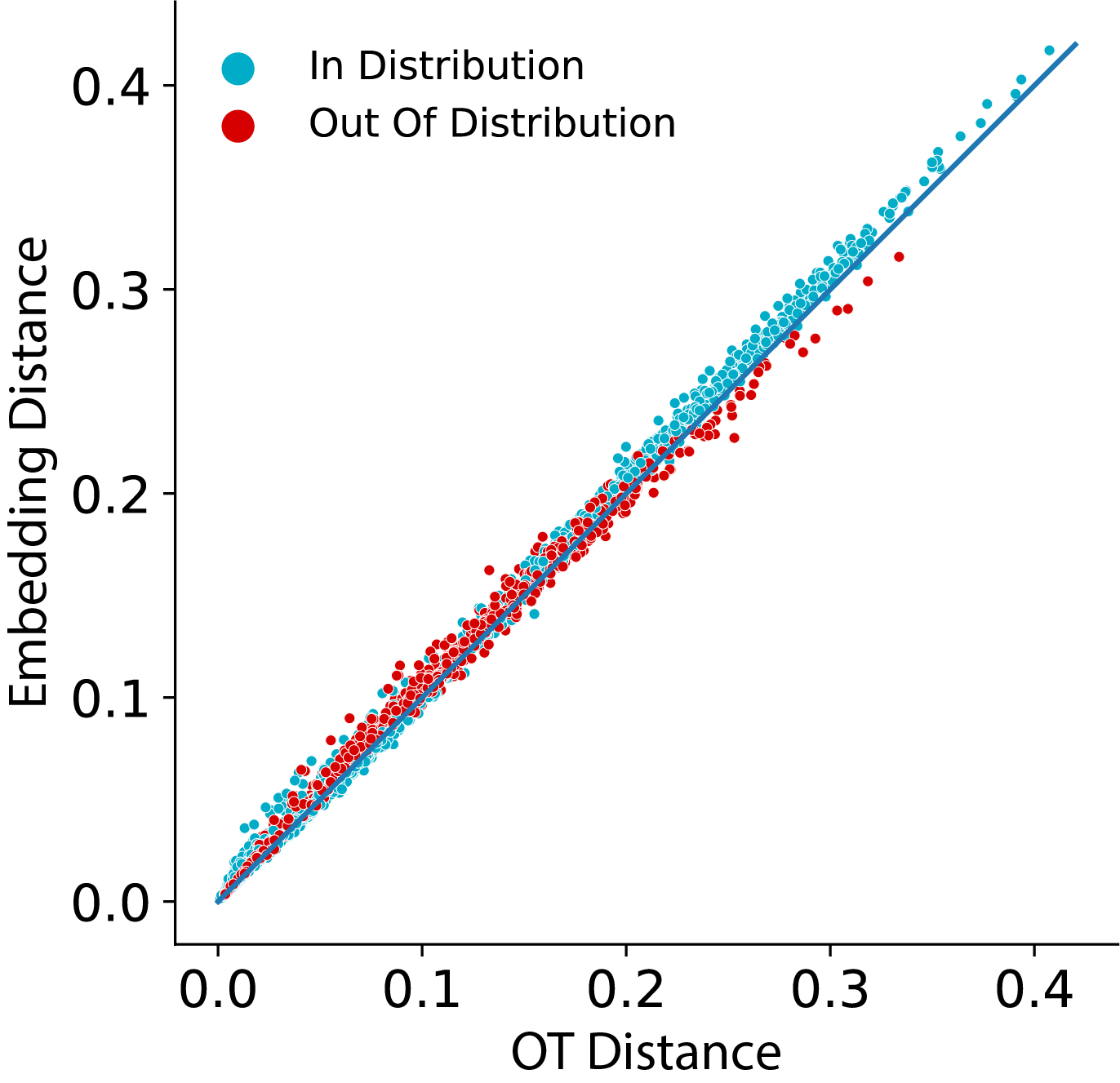}
  \caption{\textbf{Generalization to OOD on Fashion-MNIST}. During training, we held out out point clouds labeled 'Bag'. Performance was slightly lower compared to observed classes, especially for larger Wasserstein distances, as the MSE for OOD samples was $4.38\cdot10^{-5}$ as opposed to $3.37\cdot10^{-5}$ for observed samples. While not trained on any 'Bag' point cloud, their encodings still largely agree with true OT distance, producing a Pearson correlation of $\rho=0.995$ and label accuracy of $0.97$.}
  \label{mainfig:FashionMNIST}
\vspace{-1em}
\end{figure}

\subsection{MNIST (Touchstone)}

We applied Wormhole to the MNIST dataset of hand-written digits, which consists of 70,000 samples of $28 \times 28$ pixel grayscale images. We converted each sample to a point cloud by thresholding and selecting the set of pixel coordinates that passed the threshold (Figure \ref{mainfig:MNIST}a). While MNIST is considered small by modern machine-learning standards, straightforward computation of all $\binom{70000}{2}= 2.45\cdot10^9$ pairwise Wasserstein distances using current OT solvers is far from computationally feasible and would require a week of JAX-OTT \cite{cuturi2022optimal} on an 80GB GPU (Figure \ref{mainfig:time_benchmark}).

By contrast, training Wormhole with default parameters (Supplementary Section \ref{section:imp_details}) requires a small fraction of OT computations (approximately a $2000$-fold reduction) compared to the naive approach, with a total runtime of only 10 min. Despite the speedup, Wormhole still achieves remarkable performance, learning common patterns within the MNIST dataset and significantly outperforming contemporary approaches. Pairwise distances between point cloud encodings match their true OT distances and represent underlying digits (Figure \ref{mainfig:MNIST}b,c).

\subsection{Fashion-MNIST (Out-Of-Distribution)}

Similar to MNIST, FashionMNIST contains 70,000 images that each measure $28\times28$ pixels, but the images depict articles of clothing instead of digits. After thresholding FashionMNIST images to produce point clouds, Wormhole was able to accurately recover OT distances and sample labels (Table \ref{table:summary}) in the embedded space.

For a model to be deployed confidently, it is imperative that it can generalize to out-of-distribution (OOD) examples. To evaluate the capabilities of Wormhole on OOD data, we hid all examples labeled 'Bag' during training and measured the quality of encodings. We found that Wormhole preserves the OT metric (Figure \ref{mainfig:FashionMNIST}). In addition, when classifying test-set encodings based on in-distribution and OOD samples, accuracy does not diminish on the held-out class. Regardless of whether or not 'Bag' samples were included in Wormhole training, their test-set accuracy was consistently $0.97$. We repeat this experiment for all $10$ classes in the FashionMNIST cohort and find that results are robust, having an overall OOD error of $3.14\cdot10^-5\pm2.22\cdot10^{-5}$ and correlation of $0.998\pm0.001$.

\subsection{ModelNet40 (Variational Wasserstein)}

The ModelNet40 dataset comprises point clouds from 3D synthetic CAD models of 40 diverse object classes. While each point cloud is more complex than the image datasets, containing thousands of points as opposed to hundreds, Wormhole was able to produce accurate embeddings in $30$ GPU min of training time (Table \ref{table:summary}). 

\begin{figure}[h]
\centering
\includegraphics[width=0.45\textwidth]{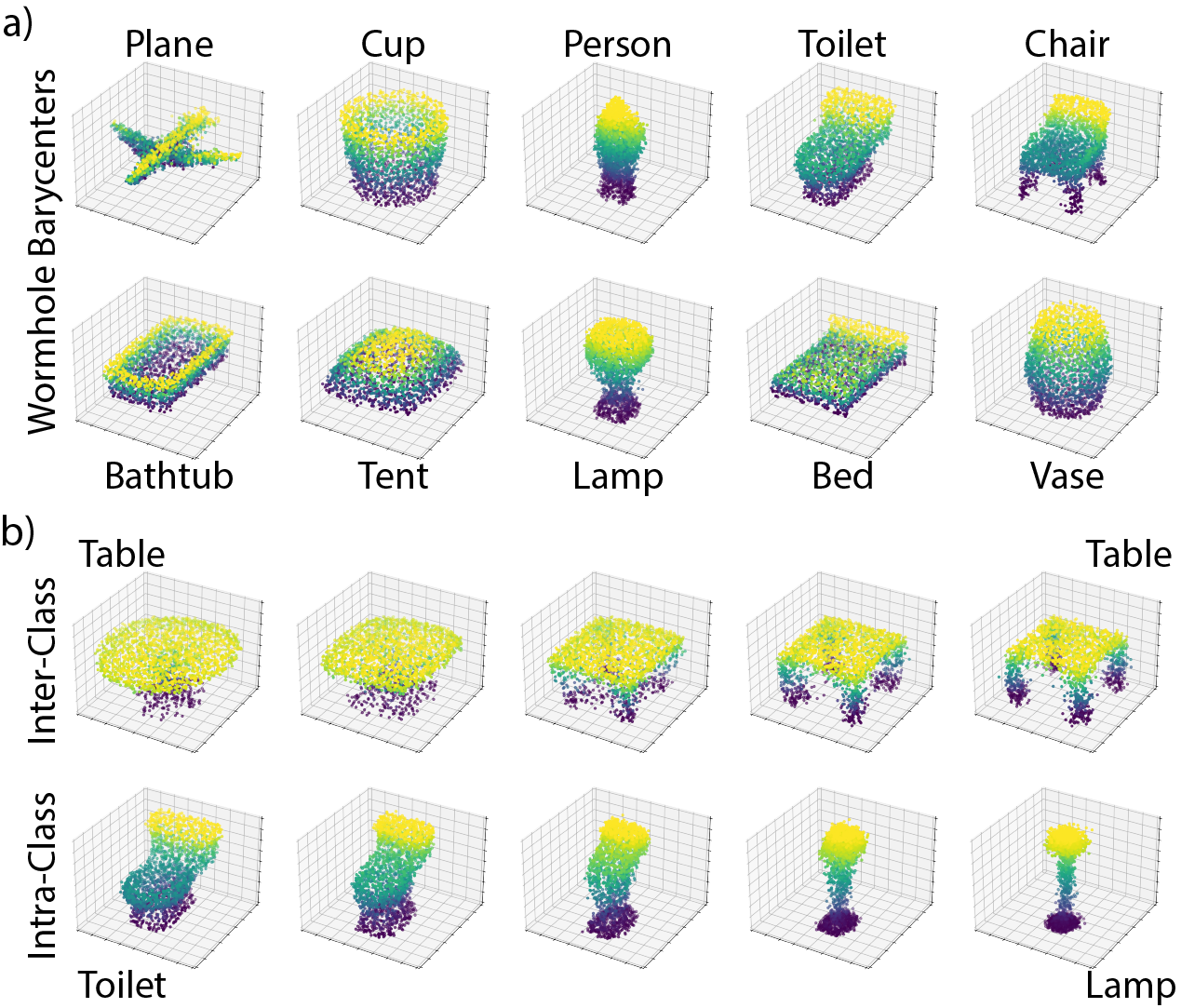}
\caption{\textbf{Wormhole Variational Wasserstein. a.} Mean encodings of each class were passed through the decoder to visualize their Wasserstein barycenters. Working in Wormhole space provides a quick and straightforward proxy for OT analysis, and recovered decodings reproduce the underlying class well. For brevity, we show 10 out of the 40 classes in from ModelNet40. \textbf{b.} Wormhole interpolation between two point clouds within the same class (top) and across different classes (bottom). Intermediate point clouds are decodings of linear interpolations between embeddings.}
\label{mainfig:Variational}
\end{figure}



As a more challenging test case, we assessed the ability of the Wormhole decoder to interpret the embeddings and solve variational Wasserstein problems. The Wasserstein barycenters for a set of distributions is their average OT distance. Calculating barycenters is computationally challenging even in a small cohort. Wormhole embedding space is a Euclidean approximation of Wasserstein and barycenters of a set can be computed by calculating their encoding average and decoding it (Figure \ref{mainfig:Variational}a).

Using Wormhole, we can also smoothly transition between point clouds via Wasserstein interpolation in the embedded space. Interpolating between distributions in Wasserstein space involves solving a weighted version of the barycenter problem as many times as there are intermediate samples. With Wormhole, we can leverage the Euclidean approximation of its latent space and directly decode linear interpolations of embeddings (Figure \ref{mainfig:Variational}b). Despite interpolated embeddings lying outside of  Wormhole training distribution, their decodings reliably visualize point cloud transitions. 


\subsection{ShapeNet (Gromov-Wasserstein Wormhole)}

Gromov-Wasserstein (GW) \cite{memoli2011gromov} is an extension of OT which seeks the most isometric map between point clouds. Given two point clouds, their GW distance is:

\begin{gather}\label{GW}
  GW(X, Y) = \min_{P\in U(\nu,\mu)}\sum_{i,j,i',j'}\lVert d^{X}_{i,i'}-d^{Y}_{j,j'} \rVert^{2}P_{i,j}P_{i',j'}
\end{gather}

Where $d^{X}_{i,i'}$ is the distance between $X_{i}$ and $X_{i'}$, and similarly for $d^{Y}_{j,j'}$. This problem is quadratic is significantly more computationally intensive than standard OT \cite{peyre2019computational}. Unlike standard OT, GW can align datasets from different modalities \cite{demetci2020gromov}, map between graphs \cite{xu2019gromov} and quantify distances between 3D shapes with arbitrary orientation \cite{memoli2009spectral}. 

\begin{figure}[h]
  \centering
  \includegraphics[width=0.45\textwidth]{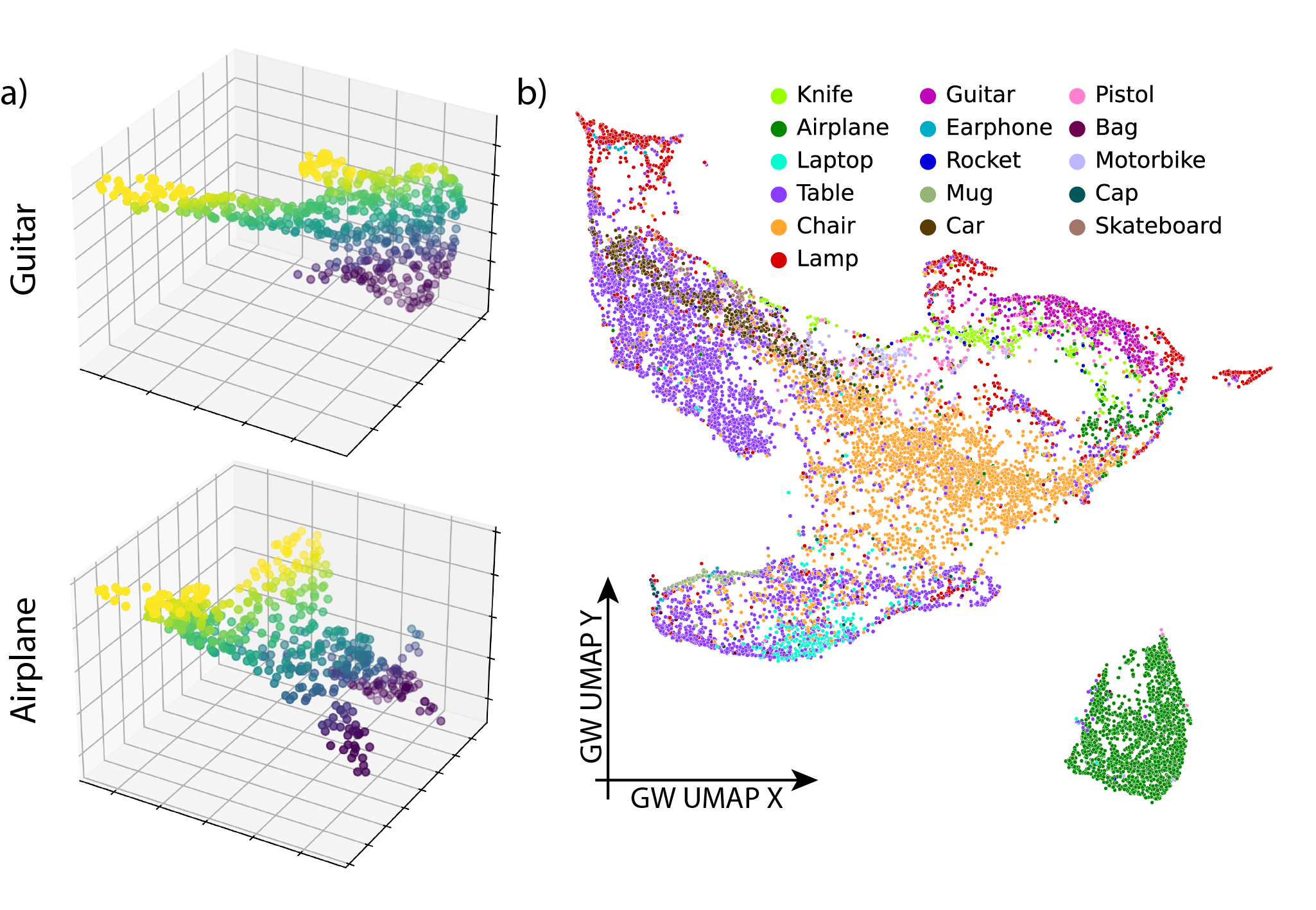}
  \caption{\textbf{Gromov-Wasserstein Wormhole. a.} Examples of point clouds from the ShapeNet dataset randomly rotated in 3D to reflect common GW applications. \textbf{b.} Embeddings learned by GW Wormhole, visualized in 2D with UMAP and colored by object class. Despite their random orientations, Wormhole was able to encode the ShapeNet point clouds and represent their class, as well as recover GW distances (Table \ref{table:summary}).}
\label{mainfig:GW}
\vspace{-1em}
\end{figure}

ShapeNet is a CAD-based dataset of 3D point clouds consisting of 16 different object classes. To construct the conditions where GW is usually applied, we randomly rotated each sample in 3D dimensions (Figure \ref{mainfig:GW}a). Standard OT is blind to structure when orientation is unknown, making rotation-invariant GW necessary. We then trained Wormhole to encode the rotated point clouds such that distances between embeddings retain GW distances.

Although the training objective was radically different, Wormhole generalized to GW and learned accurate embeddings (Table \ref{table:summary}). In addition, the encodings were reflective of object classes despite input coordinates being randomly rotated, implying that GW Wormhole was able to produce rotation invariant embeddings (Figure \ref{mainfig:GW}b).

\subsection{Cellular Niches (High-Dimensional Data)}\label{subsection:niches}

Spatially resolved transcriptomics (ST) are a set of technologies which capture both RNA information and the spatial context of cells to gain insight into tissue organization. OT has been frequently applied to analyze results of ST and unearth the relationship between a cell's phenotype and its microenvironment \cite{cang2020inferring, haviv2024covariance}. For example, Spatial Optimal Transport (SPOT) \cite{mani2022spot} models the niche of each cell as the point cloud from the gene expression of every cell in its environment, and defines differences between niches using Wasserstein.

While OT is an alluring metric to quantify cellular niches, SPOT cannot scale to ST datasets of typical size. As experimental technologies mature, atlas-level spatial assays continue to grow in ubiquity and size. For example, the $254$-gene MERFISH dataset of the motor cortex \cite{zhang2021spatially}, which consists of 256,058 cells across 64 slices. 

We summarize the $50\mu m$ radius around each cell in the MERFISH dataset to construct niche point clouds in $254$ dimensions. OT linearization methods such as DWE and DiffusionEMD assume that data is voxelized or discretized. While low-dimensional datasets can be manipulated to fit within their frameworks, the curse of dimensionality prevents adequate voxelization of this \textit{Cell Niches} dataset. As commercial technologies capture more genes \cite{marco2023optimizing}, the need for a method that can scale to higher dimensions, such as Wormhole, becomes more acute. 

\begin{figure}[h]
  \centering
  \includegraphics[width=0.45\textwidth]{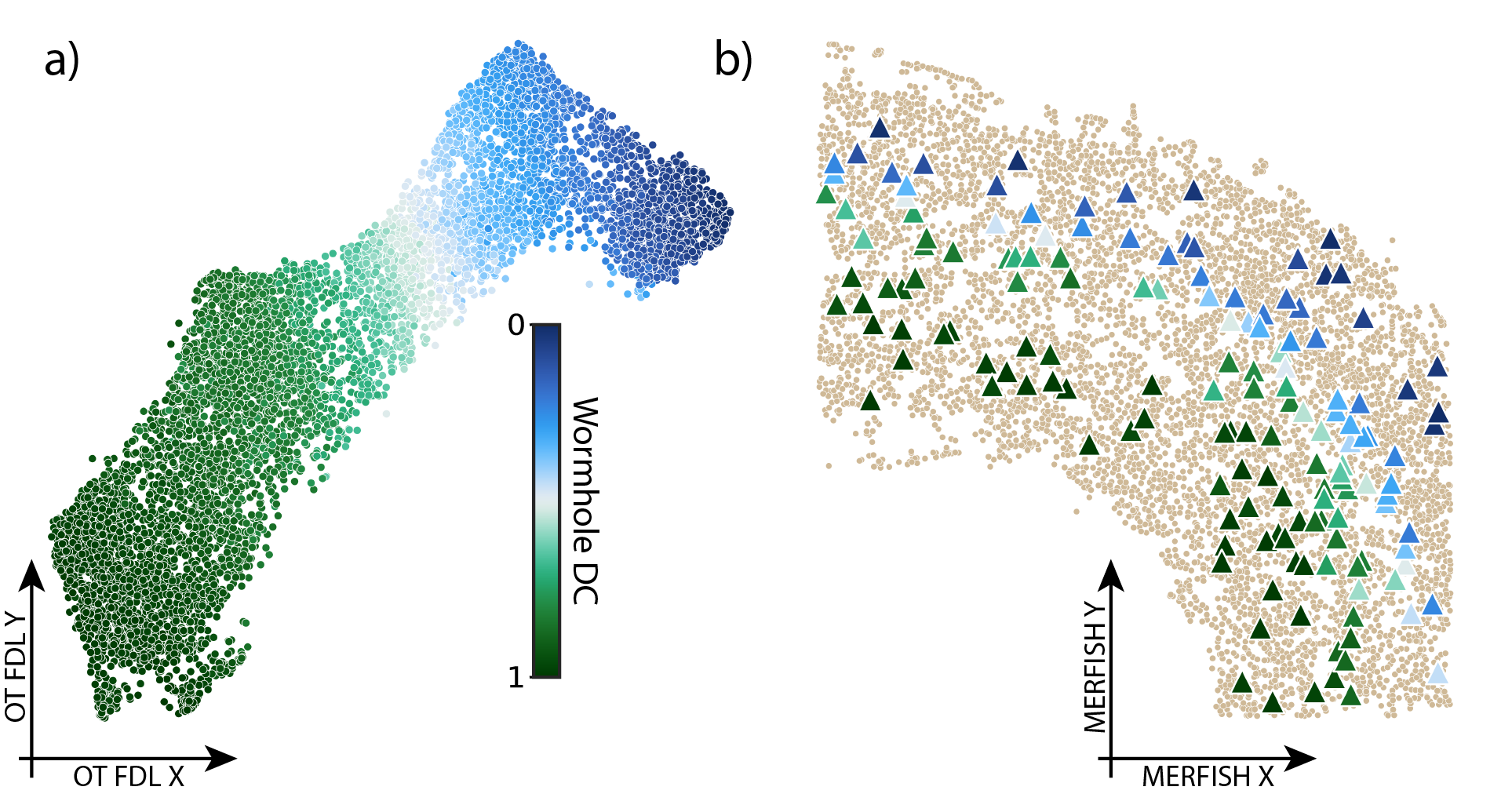}
  \caption{\textbf{Wasserstein Wormhole on 254 dimensional spatial transcriptomics data. a.} Visualization of Wormhole embeddings from niches of \textit{Pvalb} neurons from the MERFISH dataset, colored by their first diffusion component. \textbf{b.} Spatial organization of \textit{Pvalb} cells in a representative MERFISH slice. Wormhole embeddings reveal the cortical arrangement of these cells, as the first DC corresponds to neuronal depth.}
  \label{mainfig:MERFISH}
\vspace{-1em}
\end{figure}

Parvalbumin-positive (\textit{Pvalb}) interneurons are one the major groups of inhibitory neuron in the cortex, and are crucial for the regulation of neural circuitry \cite{sohal2009parvalbumin}. They are sparse in the motor cortex, comprising only $3\%$ of the dataset. To understand their spatial organization, it is therefore critical to study the entire cohort and integrate insights from multiple tissue slices.

After training Wormhole on the complete datasets (Table \ref{table:summary}), we analyze the embeddings of the niches of \textit{Pvalb} neurons. The first diffusion component (DC) \cite{coifman2006diffusion} of Wormhole encodings aligns with their cortical depth across all slices within the dataset (Figure \ref{mainfig:MERFISH}a-b). Despite the size and dimensionality of this dataset, Wormhole can still identify key biological signals across the cohort. 

\section{Discussion}

We have introduced Wasserstein Wormhole, an approach for embedding point clouds such that OT distances are preserved and can be approximated efficiently. Although OT is ubiquitous, applying it to cohorts of distributions requires costly computations and scales quadratically with cohort size. Wormhole expands OT to large datasets.

Encoding based on OT distances is theoretically challenged by the fact that Wasserstein is a non-Euclidean metric. To alleviate this concern, we derived theoretical lower and upper bounds that delimit how far non-Euclidean distance matrices can be from the embeddings. We further describe an algorithm based on projected gradient descent which is guaranteed to find an optimal embedding. On examples where the full Wasserstein distance matrix is tractable, Wormhole falls within the bounds and is close to the optimal error.

We applied our approach to eight different settings, ranging from 2D point clouds to massive spatial transcriptomics datasets with hundreds of genes. Across all settings, Wormhole embeddings recapitulated the Wasserstein space, accurately encoded key aspects of the point clouds and outperformed competing approaches. Furthermore, Wormhole's decoder recovered solutions to variational Wasserstein problems such as barycenters and point cloud interpolations. Wormhole seamlessly extends to other OT metrics, such as GW, and learns from arbitrary orientations.

Training Wasserstein Wormhole never exceeded 30 min on a single GPU, even on the 2048 points-per-sample ModelNet40 \& ShapeNet datasets. Our algorithm is implemented in JAX \cite{bradbury2021jax} and integrates with OT tools (OTT-JAX) \cite{cuturi2022optimal} for efficient Wasserstein distance calculations. Wormhole can be easily applied to any point cloud dataset via a \textit{pip} installable package.

Several methods have been proposed to learn approximations for the Wasserstein distance, including those based on neural networks. Unlike prior methods, however, Wormhole makes no assumptions about empirical distributions and can be applied to general point clouds in arbitrary dimensions, while providing superior performance. 

We envision several future directions for this work. Currently, Wormhole is only equipped to handle uniform point-clouds, where each point is equally weighted. Future work can extend the Attention mechanism to include arbitrarily weighted point-clouds, to reflect more general instances of empirical distributions and applications of OT.

Wasserstein is not only used to define a distance between distributions, but also to find the optimal mapping across them. An alluring Wormhole extension would be for it to produce the ideal matching in addition to estimating distance. This could possibly be achieved by exploiting the differentiability of Wormhole embeddings with respect to the point clouds.

\section*{Acknowledgements}

This work was supported by National Cancer Institute grants P30 CA08748 and U54 CA209975. Dana Pe'er is an Investigator at the Howard Hughes Medical Institute.

\section*{Impact Statement}

Wasserstein Wormhole is a novel method to scale computations of OT distances to large cohorts. We currently do not foresee any adverse effects of our work

\bibliography{main.bib} 

\begin{thebibliography}{64}
\providecommand{\natexlab}[1]{#1}
\providecommand{\url}[1]{\texttt{#1}}
\expandafter\ifx\csname urlstyle\endcsname\relax
  \providecommand{\doi}[1]{doi: #1}\else
  \providecommand{\doi}{doi: \begingroup \urlstyle{rm}\Url}\fi

\bibitem[Amos et~al.(2022)Amos, Cohen, Luise, and Redko]{amos2022meta}
Amos, B., Cohen, S., Luise, G., and Redko, I.
\newblock Meta optimal transport.
\newblock \emph{arXiv preprint arXiv:2206.05262}, 2022.

\bibitem[Bachmann et~al.(2022)Bachmann, Hennig, and Kobak]{bachmann2022wasserstein}
Bachmann, F., Hennig, P., and Kobak, D.
\newblock Wasserstein t-sne.
\newblock In \emph{Joint European Conference on Machine Learning and Knowledge Discovery in Databases}, pp.\  104--120. Springer, 2022.

\bibitem[Boyd \& Vandenberghe(2004)Boyd and Vandenberghe]{boyd2004convex}
Boyd, S.~P. and Vandenberghe, L.
\newblock \emph{Convex optimization}.
\newblock Cambridge university press, 2004.

\bibitem[Boyle \& Dykstra(1986)Boyle and Dykstra]{boyle1986method}
Boyle, J.~P. and Dykstra, R.~L.
\newblock A method for finding projections onto the intersection of convex sets in hilbert spaces.
\newblock In \emph{Advances in Order Restricted Statistical Inference: Proceedings of the Symposium on Order Restricted Statistical Inference held in Iowa City, Iowa, September 11--13, 1985}, pp.\  28--47. Springer, 1986.

\bibitem[Bradbury et~al.(2021)Bradbury, Frostig, Hawkins, Johnson, Leary, Maclaurin, Necula, Paszke, VanderPlas, Wanderman-Milne, et~al.]{bradbury2021jax}
Bradbury, J., Frostig, R., Hawkins, P., Johnson, M.~J., Leary, C., Maclaurin, D., Necula, G., Paszke, A., VanderPlas, J., Wanderman-Milne, S., et~al.
\newblock Jax: Autograd and xla.
\newblock \emph{Astrophysics Source Code Library}, pp.\  ascl--2111, 2021.

\bibitem[Bunne et~al.(2023)Bunne, Stark, Gut, Del~Castillo, Levesque, Lehmann, Pelkmans, Krause, and R{\"a}tsch]{bunne2023learning}
Bunne, C., Stark, S.~G., Gut, G., Del~Castillo, J.~S., Levesque, M., Lehmann, K.-V., Pelkmans, L., Krause, A., and R{\"a}tsch, G.
\newblock Learning single-cell perturbation responses using neural optimal transport.
\newblock \emph{Nature Methods}, 20\penalty0 (11):\penalty0 1759--1768, 2023.

\bibitem[Cang \& Nie(2020)Cang and Nie]{cang2020inferring}
Cang, Z. and Nie, Q.
\newblock Inferring spatial and signaling relationships between cells from single cell transcriptomic data.
\newblock \emph{Nature communications}, 11\penalty0 (1):\penalty0 2084, 2020.

\bibitem[Chang et~al.(2015)Chang, Funkhouser, Guibas, Hanrahan, Huang, Li, Savarese, Savva, Song, Su, et~al.]{chang2015shapenet}
Chang, A.~X., Funkhouser, T., Guibas, L., Hanrahan, P., Huang, Q., Li, Z., Savarese, S., Savva, M., Song, S., Su, H., et~al.
\newblock Shapenet: An information-rich 3d model repository.
\newblock \emph{arXiv preprint arXiv:1512.03012}, 2015.

\bibitem[Chewi et~al.(2021)Chewi, Clancy, Le~Gouic, Rigollet, Stepaniants, and Stromme]{chewi2021fast}
Chewi, S., Clancy, J., Le~Gouic, T., Rigollet, P., Stepaniants, G., and Stromme, A.
\newblock Fast and smooth interpolation on wasserstein space.
\newblock In \emph{International Conference on Artificial Intelligence and Statistics}, pp.\  3061--3069. PMLR, 2021.

\bibitem[Coifman \& Lafon(2006)Coifman and Lafon]{coifman2006diffusion}
Coifman, R.~R. and Lafon, S.
\newblock Diffusion maps.
\newblock \emph{Applied and computational harmonic analysis}, 21\penalty0 (1):\penalty0 5--30, 2006.

\bibitem[Courty et~al.(2017)Courty, Flamary, and Ducoffe]{courty2017learning}
Courty, N., Flamary, R., and Ducoffe, M.
\newblock Learning wasserstein embeddings.
\newblock \emph{arXiv preprint arXiv:1710.07457}, 2017.

\bibitem[Cuturi(2013)]{cuturi2013sinkhorn}
Cuturi, M.
\newblock Sinkhorn distances: Lightspeed computation of optimal transport.
\newblock \emph{Advances in neural information processing systems}, 26, 2013.

\bibitem[Cuturi \& Doucet(2014)Cuturi and Doucet]{cuturi2014fast}
Cuturi, M. and Doucet, A.
\newblock Fast computation of wasserstein barycenters.
\newblock In \emph{International conference on machine learning}, pp.\  685--693. PMLR, 2014.

\bibitem[Cuturi et~al.(2022)Cuturi, Meng-Papaxanthos, Tian, Bunne, Davis, and Teboul]{cuturi2022optimal}
Cuturi, M., Meng-Papaxanthos, L., Tian, Y., Bunne, C., Davis, G., and Teboul, O.
\newblock Optimal transport tools (ott): A jax toolbox for all things wasserstein.
\newblock \emph{arXiv preprint arXiv:2201.12324}, 2022.

\bibitem[De~Leeuw \& Mair(2009)De~Leeuw and Mair]{de2009multidimensional}
De~Leeuw, J. and Mair, P.
\newblock Multidimensional scaling using majorization: Smacof in r.
\newblock \emph{Journal of statistical software}, 31:\penalty0 1--30, 2009.

\bibitem[Delon et~al.(2022)Delon, Desolneux, and Salmona]{delon2022gromov}
Delon, J., Desolneux, A., and Salmona, A.
\newblock Gromov--wasserstein distances between gaussian distributions.
\newblock \emph{Journal of Applied Probability}, 59\penalty0 (4):\penalty0 1178--1198, 2022.

\bibitem[Demetci et~al.(2020)Demetci, Santorella, Sandstede, Noble, and Singh]{demetci2020gromov}
Demetci, P., Santorella, R., Sandstede, B., Noble, W.~S., and Singh, R.
\newblock Gromov-wasserstein optimal transport to align single-cell multi-omics data.
\newblock \emph{BioRxiv}, pp.\  2020--04, 2020.

\bibitem[Dowson \& Landau(1982)Dowson and Landau]{dowson1982frechet}
Dowson, D. and Landau, B.
\newblock The fr{\'e}chet distance between multivariate normal distributions.
\newblock \emph{Journal of multivariate analysis}, 12\penalty0 (3):\penalty0 450--455, 1982.

\bibitem[Feydy et~al.(2017)Feydy, Charlier, Vialard, and Peyr{\'e}]{feydy2017optimal}
Feydy, J., Charlier, B., Vialard, F.-X., and Peyr{\'e}, G.
\newblock Optimal transport for diffeomorphic registration.
\newblock In \emph{Medical Image Computing and Computer Assisted Intervention- MICCAI 2017: 20th International Conference, Quebec City, QC, Canada, September 11-13, 2017, Proceedings, Part I 20}, pp.\  291--299. Springer, 2017.

\bibitem[Flamary et~al.(2021)Flamary, Courty, Gramfort, Alaya, Boisbunon, Chambon, Chapel, Corenflos, Fatras, Fournier, et~al.]{flamary2021pot}
Flamary, R., Courty, N., Gramfort, A., Alaya, M.~Z., Boisbunon, A., Chambon, S., Chapel, L., Corenflos, A., Fatras, K., Fournier, N., et~al.
\newblock Pot: Python optimal transport.
\newblock \emph{The Journal of Machine Learning Research}, 22\penalty0 (1):\penalty0 3571--3578, 2021.

\bibitem[Frogner et~al.(2019)Frogner, Mirzazadeh, and Solomon]{frogner2019learning}
Frogner, C., Mirzazadeh, F., and Solomon, J.
\newblock Learning embeddings into entropic wasserstein spaces.
\newblock \emph{arXiv preprint arXiv:1905.03329}, 2019.

\bibitem[Gaffke \& Mathar(1989)Gaffke and Mathar]{gaffke1989cyclic}
Gaffke, N. and Mathar, R.
\newblock A cyclic projection algorithm via duality.
\newblock \emph{Metrika}, 36\penalty0 (1):\penalty0 29--54, 1989.

\bibitem[Garcia-Garcia et~al.(2016)Garcia-Garcia, Gomez-Donoso, Garcia-Rodriguez, Orts-Escolano, Cazorla, and Azorin-Lopez]{garcia2016pointnet}
Garcia-Garcia, A., Gomez-Donoso, F., Garcia-Rodriguez, J., Orts-Escolano, S., Cazorla, M., and Azorin-Lopez, J.
\newblock Pointnet: A 3d convolutional neural network for real-time object class recognition.
\newblock In \emph{2016 International joint conference on neural networks (IJCNN)}, pp.\  1578--1584. IEEE, 2016.

\bibitem[Genevay et~al.(2018)Genevay, Peyr{\'e}, and Cuturi]{genevay2018learning}
Genevay, A., Peyr{\'e}, G., and Cuturi, M.
\newblock Learning generative models with sinkhorn divergences.
\newblock In \emph{International Conference on Artificial Intelligence and Statistics}, pp.\  1608--1617. PMLR, 2018.

\bibitem[Gower(1985)]{GOWER198581}
Gower, J.
\newblock Properties of euclidean and non-euclidean distance matrices.
\newblock \emph{Linear Algebra and its Applications}, 67:\penalty0 81--97, 1985.
\newblock ISSN 0024-3795.
\newblock \doi{https://doi.org/10.1016/0024-3795(85)90187-9}.
\newblock URL \url{https://www.sciencedirect.com/science/article/pii/0024379585901879}.

\bibitem[Guo et~al.(2020)Guo, Wang, Hu, Liu, Liu, and Bennamoun]{guo2020deep}
Guo, Y., Wang, H., Hu, Q., Liu, H., Liu, L., and Bennamoun, M.
\newblock Deep learning for 3d point clouds: A survey.
\newblock \emph{IEEE transactions on pattern analysis and machine intelligence}, 43\penalty0 (12):\penalty0 4338--4364, 2020.

\bibitem[Haviv et~al.(2023)Haviv, Gatie, Hadjantonakis, Nawy, and Pe’er]{haviv2023covariance}
Haviv, D., Gatie, M., Hadjantonakis, A.-K., Nawy, T., and Pe’er, D.
\newblock The covariance environment defines cellular niches for spatial inference.
\newblock \emph{bioRxiv}, 2023.

\bibitem[Haviv et~al.(2024)Haviv, Rem{\v{s}}{\'\i}k, Gatie, Snopkowski, Takizawa, Pereira, Bashkin, Jovanovich, Nawy, Chaligne, et~al.]{haviv2024covariance}
Haviv, D., Rem{\v{s}}{\'\i}k, J., Gatie, M., Snopkowski, C., Takizawa, M., Pereira, N., Bashkin, J., Jovanovich, S., Nawy, T., Chaligne, R., et~al.
\newblock The covariance environment defines cellular niches for spatial inference.
\newblock \emph{Nature Biotechnology}, pp.\  1--12, 2024.

\bibitem[Higham(1988)]{higham1988computing}
Higham, N.~J.
\newblock Computing a nearest symmetric positive semidefinite matrix.
\newblock \emph{Linear algebra and its applications}, 103:\penalty0 103--118, 1988.

\bibitem[Horn \& Johnson(2012)Horn and Johnson]{horn2012matrix}
Horn, R.~A. and Johnson, C.~R.
\newblock \emph{Matrix analysis}.
\newblock Cambridge university press, 2012.

\bibitem[Katageri et~al.(2024)Katageri, Sarkar, and Sharma]{katageri2024metric}
Katageri, S., Sarkar, S., and Sharma, C.
\newblock Metric learning for 3d point clouds using optimal transport.
\newblock In \emph{Proceedings of the IEEE/CVF Winter Conference on Applications of Computer Vision}, pp.\  561--569, 2024.

\bibitem[Kingma \& Ba(2014)Kingma and Ba]{kingma2014adam}
Kingma, D.~P. and Ba, J.
\newblock Adam: A method for stochastic optimization.
\newblock \emph{arXiv preprint arXiv:1412.6980}, 2014.

\bibitem[Kruskal \& Wish(1978)Kruskal and Wish]{kruskal1978multidimensional}
Kruskal, J.~B. and Wish, M.
\newblock \emph{Multidimensional scaling}.
\newblock Number~11. Sage, 1978.

\bibitem[Lee et~al.(2019)Lee, Lee, Kim, Kosiorek, Choi, and Teh]{lee2019set}
Lee, J., Lee, Y., Kim, J., Kosiorek, A., Choi, S., and Teh, Y.~W.
\newblock Set transformer: A framework for attention-based permutation-invariant neural networks.
\newblock In \emph{International conference on machine learning}, pp.\  3744--3753. PMLR, 2019.

\bibitem[Liu et~al.(2022)Liu, Bai, Lu, Soltoggio, and Kolouri]{liu2022wasserstein}
Liu, X., Bai, Y., Lu, Y., Soltoggio, A., and Kolouri, S.
\newblock Wasserstein task embedding for measuring task similarities.
\newblock \emph{arXiv preprint arXiv:2208.11726}, 2022.

\bibitem[Lohoff et~al.(2022)Lohoff, Ghazanfar, Missarova, Koulena, Pierson, Griffiths, Bardot, Eng, Tyser, Argelaguet, et~al.]{lohoff2022integration}
Lohoff, T., Ghazanfar, S., Missarova, A., Koulena, N., Pierson, N., Griffiths, J., Bardot, E., Eng, C.-H., Tyser, R., Argelaguet, R., et~al.
\newblock Integration of spatial and single-cell transcriptomic data elucidates mouse organogenesis.
\newblock \emph{Nature biotechnology}, 40\penalty0 (1):\penalty0 74--85, 2022.

\bibitem[Mani et~al.(2022)Mani, Haviv, Kunes, and Pe'er]{mani2022spot}
Mani, S., Haviv, D., Kunes, R., and Pe'er, D.
\newblock Spot: Spatial optimal transport for analyzing cellular microenvironments.
\newblock In \emph{NeurIPS 2022 Workshop on Learning Meaningful Representations of Life}, 2022.

\bibitem[Marco~Salas et~al.(2023)Marco~Salas, Czarnewski, Kuemmerle, Helgadottir, Mattsson~Langseth, Tiesmeyer, Avenel, Rehman, Tiklova, Andersson, et~al.]{marco2023optimizing}
Marco~Salas, S., Czarnewski, P., Kuemmerle, L.~B., Helgadottir, S., Mattsson~Langseth, C., Tiesmeyer, S., Avenel, C., Rehman, H., Tiklova, K., Andersson, A., et~al.
\newblock Optimizing xenium in situ data utility by quality assessment and best practice analysis workflows.
\newblock \emph{bioRxiv}, pp.\  2023--02, 2023.

\bibitem[M{\'e}moli(2009)]{memoli2009spectral}
M{\'e}moli, F.
\newblock Spectral gromov-wasserstein distances for shape matching.
\newblock In \emph{2009 IEEE 12th International Conference on Computer Vision Workshops, ICCV Workshops}, pp.\  256--263. IEEE, 2009.

\bibitem[M{\'e}moli(2011)]{memoli2011gromov}
M{\'e}moli, F.
\newblock Gromov--wasserstein distances and the metric approach to object matching.
\newblock \emph{Foundations of computational mathematics}, 11:\penalty0 417--487, 2011.

\bibitem[Naor \& Schechtman(2007)Naor and Schechtman]{naor2007planar}
Naor, A. and Schechtman, G.
\newblock Planar earthmover is not in l\_1.
\newblock \emph{SIAM Journal on Computing}, 37\penalty0 (3):\penalty0 804--826, 2007.

\bibitem[Nitzan et~al.(2019)Nitzan, Karaiskos, Friedman, and Rajewsky]{nitzan2019gene}
Nitzan, M., Karaiskos, N., Friedman, N., and Rajewsky, N.
\newblock Gene expression cartography.
\newblock \emph{Nature}, 576\penalty0 (7785):\penalty0 132--137, 2019.

\bibitem[Pedregosa et~al.(2011)Pedregosa, Varoquaux, Gramfort, Michel, Thirion, Grisel, Blondel, Prettenhofer, Weiss, Dubourg, et~al.]{pedregosa2011scikit}
Pedregosa, F., Varoquaux, G., Gramfort, A., Michel, V., Thirion, B., Grisel, O., Blondel, M., Prettenhofer, P., Weiss, R., Dubourg, V., et~al.
\newblock Scikit-learn: Machine learning in python.
\newblock \emph{the Journal of machine Learning research}, 12:\penalty0 2825--2830, 2011.

\bibitem[Persad et~al.(2023)Persad, Choo, Dien, Sohail, Masilionis, Chalign{\'e}, Nawy, Brown, Sharma, Pe’er, et~al.]{persad2023seacells}
Persad, S., Choo, Z.-N., Dien, C., Sohail, N., Masilionis, I., Chalign{\'e}, R., Nawy, T., Brown, C.~C., Sharma, R., Pe’er, I., et~al.
\newblock Seacells infers transcriptional and epigenomic cellular states from single-cell genomics data.
\newblock \emph{Nature Biotechnology}, 41\penalty0 (12):\penalty0 1746--1757, 2023.

\bibitem[Peyr{\'e} et~al.(2019)Peyr{\'e}, Cuturi, et~al.]{peyre2019computational}
Peyr{\'e}, G., Cuturi, M., et~al.
\newblock Computational optimal transport: With applications to data science.
\newblock \emph{Foundations and Trends{\textregistered} in Machine Learning}, 11\penalty0 (5-6):\penalty0 355--607, 2019.

\bibitem[Qi et~al.(2017)Qi, Su, Mo, and Guibas]{qi2017pointnet}
Qi, C.~R., Su, H., Mo, K., and Guibas, L.~J.
\newblock Pointnet: Deep learning on point sets for 3d classification and segmentation.
\newblock In \emph{Proceedings of the IEEE conference on computer vision and pattern recognition}, pp.\  652--660, 2017.

\bibitem[Scetbon et~al.(2021)Scetbon, Cuturi, and Peyr{\'e}]{scetbon2021low}
Scetbon, M., Cuturi, M., and Peyr{\'e}, G.
\newblock Low-rank sinkhorn factorization.
\newblock In \emph{International Conference on Machine Learning}, pp.\  9344--9354. PMLR, 2021.

\bibitem[Schiebinger et~al.(2019)Schiebinger, Shu, Tabaka, Cleary, Subramanian, Solomon, Gould, Liu, Lin, Berube, et~al.]{schiebinger2019optimal}
Schiebinger, G., Shu, J., Tabaka, M., Cleary, B., Subramanian, V., Solomon, A., Gould, J., Liu, S., Lin, S., Berube, P., et~al.
\newblock Optimal-transport analysis of single-cell gene expression identifies developmental trajectories in reprogramming.
\newblock \emph{Cell}, 176\penalty0 (4):\penalty0 928--943, 2019.

\bibitem[Sehanobish et~al.(2020)Sehanobish, Ravindra, and van Dijk]{sehanobish2020permutation}
Sehanobish, A., Ravindra, N.~G., and van Dijk, D.
\newblock Permutation invariant networks to learn wasserstein metrics.
\newblock In \emph{TDA $\{$$\backslash$\&$\}$ Beyond}, 2020.

\bibitem[Shirdhonkar \& Jacobs(2008)Shirdhonkar and Jacobs]{shirdhonkar2008approximate}
Shirdhonkar, S. and Jacobs, D.~W.
\newblock Approximate earth mover’s distance in linear time.
\newblock In \emph{2008 IEEE Conference on Computer Vision and Pattern Recognition}, pp.\  1--8. IEEE, 2008.

\bibitem[Sohal et~al.(2009)Sohal, Zhang, Yizhar, and Deisseroth]{sohal2009parvalbumin}
Sohal, V.~S., Zhang, F., Yizhar, O., and Deisseroth, K.
\newblock Parvalbumin neurons and gamma rhythms enhance cortical circuit performance.
\newblock \emph{Nature}, 459\penalty0 (7247):\penalty0 698--702, 2009.

\bibitem[Solomon et~al.(2015)Solomon, De~Goes, Peyr{\'e}, Cuturi, Butscher, Nguyen, Du, and Guibas]{solomon2015convolutional}
Solomon, J., De~Goes, F., Peyr{\'e}, G., Cuturi, M., Butscher, A., Nguyen, A., Du, T., and Guibas, L.
\newblock Convolutional wasserstein distances: Efficient optimal transportation on geometric domains.
\newblock \emph{ACM Transactions on Graphics (ToG)}, 34\penalty0 (4):\penalty0 1--11, 2015.

\bibitem[Sonthalia et~al.(2021)Sonthalia, Van~Buskirk, Raichel, and Gilbert]{sonthalia2021can}
Sonthalia, R., Van~Buskirk, G., Raichel, B., and Gilbert, A.
\newblock How can classical multidimensional scaling go wrong?
\newblock \emph{Advances in Neural Information Processing Systems}, 34:\penalty0 12304--12315, 2021.

\bibitem[Stephenson et~al.(2021)Stephenson, Reynolds, Botting, Calero-Nieto, Morgan, Tuong, Bach, Sungnak, Worlock, Yoshida, et~al.]{stephenson2021single}
Stephenson, E., Reynolds, G., Botting, R.~A., Calero-Nieto, F.~J., Morgan, M.~D., Tuong, Z.~K., Bach, K., Sungnak, W., Worlock, K.~B., Yoshida, M., et~al.
\newblock Single-cell multi-omics analysis of the immune response in covid-19.
\newblock \emph{Nature medicine}, 27\penalty0 (5):\penalty0 904--916, 2021.

\bibitem[Su et~al.(2015)Su, Wang, Shi, Zeng, Sun, Luo, and Gu]{su2015optimal}
Su, Z., Wang, Y., Shi, R., Zeng, W., Sun, J., Luo, F., and Gu, X.
\newblock Optimal mass transport for shape matching and comparison.
\newblock \emph{IEEE transactions on pattern analysis and machine intelligence}, 37\penalty0 (11):\penalty0 2246--2259, 2015.

\bibitem[Tong et~al.(2021)Tong, Huguet, Natik, MacDonald, Kuchroo, Coifman, Wolf, and Krishnaswamy]{tong2021diffusion}
Tong, A.~Y., Huguet, G., Natik, A., MacDonald, K., Kuchroo, M., Coifman, R., Wolf, G., and Krishnaswamy, S.
\newblock Diffusion earth mover’s distance and distribution embeddings.
\newblock In \emph{International Conference on Machine Learning}, pp.\  10336--10346. PMLR, 2021.

\bibitem[Torgerson(1952)]{torgerson1952multidimensional}
Torgerson, W.~S.
\newblock Multidimensional scaling: I. theory and method.
\newblock \emph{Psychometrika}, 17\penalty0 (4):\penalty0 401--419, 1952.

\bibitem[Torrence \& Compo(1998)Torrence and Compo]{torrence1998practical}
Torrence, C. and Compo, G.~P.
\newblock A practical guide to wavelet analysis.
\newblock \emph{Bulletin of the American Meteorological society}, 79\penalty0 (1):\penalty0 61--78, 1998.

\bibitem[Vaswani et~al.(2017)Vaswani, Shazeer, Parmar, Uszkoreit, Jones, Gomez, Kaiser, and Polosukhin]{vaswani2017attention}
Vaswani, A., Shazeer, N., Parmar, N., Uszkoreit, J., Jones, L., Gomez, A.~N., Kaiser, {\L}., and Polosukhin, I.
\newblock Attention is all you need.
\newblock \emph{Advances in neural information processing systems}, 30, 2017.

\bibitem[Villani et~al.(2009)]{villani2009optimal}
Villani, C. et~al.
\newblock \emph{Optimal transport: old and new}, volume 338.
\newblock Springer, 2009.

\bibitem[Wu et~al.(2023)Wu, Courty, Jin, and Li]{wu2023improving}
Wu, F., Courty, N., Jin, S., and Li, S.~Z.
\newblock Improving molecular representation learning with metric learning-enhanced optimal transport.
\newblock \emph{Patterns}, 4\penalty0 (4), 2023.

\bibitem[Wu et~al.(2015)Wu, Song, Khosla, Yu, Zhang, Tang, and Xiao]{wu20153d}
Wu, Z., Song, S., Khosla, A., Yu, F., Zhang, L., Tang, X., and Xiao, J.
\newblock 3d shapenets: A deep representation for volumetric shapes.
\newblock In \emph{Proceedings of the IEEE conference on computer vision and pattern recognition}, pp.\  1912--1920, 2015.

\bibitem[Xu et~al.(2019)Xu, Luo, Zha, and Duke]{xu2019gromov}
Xu, H., Luo, D., Zha, H., and Duke, L.~C.
\newblock Gromov-wasserstein learning for graph matching and node embedding.
\newblock In \emph{International conference on machine learning}, pp.\  6932--6941. PMLR, 2019.

\bibitem[Zhang et~al.(2021)Zhang, Eichhorn, Zingg, Yao, Cotter, Zeng, Dong, and Zhuang]{zhang2021spatially}
Zhang, M., Eichhorn, S.~W., Zingg, B., Yao, Z., Cotter, K., Zeng, H., Dong, H., and Zhuang, X.
\newblock Spatially resolved cell atlas of the mouse primary motor cortex by merfish.
\newblock \emph{Nature}, 598\penalty0 (7879):\penalty0 137--143, 2021.

\end{thebibliography}
\bibliographystyle{icml2024}

\clearpage

\setcounter{figure}{0}
\makeatletter 
\renewcommand{\thefigure}{S\@arabic\c@figure}
\makeatother


\onecolumn
\icmltitle{Wasserstein Wormhole - Supplementary Materials}

\section{Permutation Equivariance of Attention}\label{section:permutation}

The permutation equivariance feature of self-attention is a critical component of our Wormhole algorithm. This means that switching the order of points similarly switches the order of the Attention encodings. While generally known \cite{lee2019set}, due to its relevance in our work, we provide a proof of the equivariance here too. Given weights $W_{K}, W_{V}, W_{Q} \in R^{d\times d}$, the attention operator on input $X\in R^{n\times d}$ is:

\begin{equation}
  A(X) =\text{softmax}\left(\frac{XW_{Q}W_{K}^{\top}X^{\top}}{\sqrt{d}}\right)XW_{V}
\end{equation}

So when rows of $X$ are permuted via matrix $P$, the attention output is: 

\begin{equation} 
  A(PX) = \text{softmax}\left(\frac{PXW_{Q}W_{K}^{\top}X^{\top}P^{\top}}{\sqrt{d}}\right)PXW_{V} 
\end{equation} 

Since \textit{softmax} is an element wise operation, we can factor out the permutation. From the unitary property of permutation matrices ($P^{\top}=P^{-1}$):

\begin{gather}
  A(PX) = P\cdot\text{softmax}\left(\frac{XW_{Q}W_{K}^{\top}X^{\top}}{\sqrt{d}}\right)XW_{V}
\end{gather}

Which shows that attention operator is permutation equivariant.

\section{Bounds on Embedding non-Euclidean Distances}\label{section:theory_supp}

Let $D \in \mathbb{R}^{N\times N}$ by a pairwise distance matrix. Following definition \ref{EDM:def}, if $D$ is Euclidean (EDM), is can be perfectly represented as the pairwise distances between $n$ points through multi-dimensional scaling (MDS) \cite{kruskal1978multidimensional}. Since the Wasserstein distance is non-Euclidean \cite{peyre2019computational}, we are interested in how well we can approximate non-Euclidean distance matrices with pairwise distances between learned embeddings. 

Here, we consider the problem of finding the closet valid EDM to $D$:

\begin{equation}\label{eq:convex_problem}
  S = \min_{D'} \| D - D' \|_F^2
\end{equation}

with the constraints reflecting $D'$ being an EDM:

\begin{equation*}
\text{$C_{1}$: } -J\hat{D}J \text{ is PSD, $C_{2}$: } \hat{D} \text{ is hollow and $C_{3}$: } \hat{D}_{i,j}\geq0.
\end{equation*}

This problem is fully convex and has a global minimum which we call $\mathbfcal{L^{*}}$. In the following, we derive a lower and upper for $\mathbfcal{L^{*}}$. To find a lower bound, we solve a simplified problem by eschewing the hollowness and non-negativity constraints on $D'$. Then, we will reintroduce those constraints and use the solution for the simplified problem to construct an upper bound. 

\subsection{Error Lower Bound}\label{subsection:lower_bound}

Let $P = \frac{1}{N}\mathbf{1}_N\mathbf{1}_N^{\top}$. We can write the identity matrix as $I = J + P$ and turn our objective into:

\begin{gather}\label{eq:ID}
\| D - D' \|_F^2 = \| (J + P)D(J+P) - (J + P)D'(J+P) \|_F^2 
\end{gather}

Expanding the products:

\begin{gather}
  \| D - D' \|_F^2 = \| A + B \|_F^2 
\end{gather}

where $A:= JDJ - JD'J$ and $B:= PD'J + JD'P +PD'P - PDJ - JDP - PDP$. From the properties of the Frobenius norm:

\begin{gather}\label{eq:frob_norm}
\| A + B \|_F^2 = \| A \|_F^2 + \| B \|_F^2 + 2Tr(A^{\top}B)
\end{gather}

Since the values of $P$ are constant $J\cdot P = P\cdot J = 0$, and therefore the only non-zero term in $AB$ are:

\begin{gather}
  AB = J(D-D')J(D-D')P
\end{gather}

Trace of matrix product is invariant under cyclic permutation, which means:

\begin{gather}
Tr(AB) = Tr(J(D-D')J(D-D')P) = 0
\end{gather}

where again we use the identity that $P\cdot J = 0$. Plugging that back in equation \ref{eq:frob_norm}:

\begin{equation}
  \|D- D'\|^2_F = \|A\|^2_F + \|B\|^2_F
\end{equation}

Conceptually, the $\|A\|^2_F$ term is optimized when the $D'$ and $D$ are equal after centering, and therefore does not depend on their row/columns means. Meanwhile the $\|B\|^2_F$ term is optimized when the row/columns means of $D'$ and $D$ are equal. 

When only the constraint on $F'=-JD'J$ being PSD is maintained, the two terms $\|A\|^2_F$ and $\|B\|^2_F$ can be solved independently to derive an analytical solution. The simplified version of optimizing $\|A\|^2_F$ is:

\begin{equation}
\begin{gathered}
\min_{D_{A}} \| JD_{A}J - JDJ\|_F^2 \\ 
\textrm{s.t.} -JD_{A}J \; \textrm{is PSD} 
\end{gathered}
\end{equation}

This describes the problem of finding the closest PSD approximation to a symmetric matrix \cite{higham1988computing}, whose solution is truncation of the negative eigenspace of $JDJ$:

\begin{equation}\label{eq:truncation}
  D_{A}= - \sum_{i: \lambda_{i}>0}\lambda_iv_{i}v^{\top}_{i}
\end{equation}

where $\{\lambda_i, v_i\}_{i=1}^N$ are the eigenvalues and eigenvectors of $F=-JDJ$. 

Since $JDJ$ has row/column means 0, the ones vector $\mathbf{1}_{n}$ is an eigenvector of $JDJ$ with eigenvalue $0$. Therefore $\mathbf{1}_{n}$ is also in the null space $D_{A}$ and:
\begin{equation}
  D_{A} = JD_{A}J
\end{equation}

and 

\begin{equation}
  PD_{A} = D_{A}P = 0
\end{equation}

To solve $\|B\|^2_F$, we apply the definition of $J$ and arrive at:

\begin{equation}
  PDJ + JDP + PDP = PD + (PD)^{\top} - PDP
\end{equation}

We denote the mean of the $i$-th row/column of $D$ as $\hat{D_{i}}$ and the mean of all elements as $\hat{D}$. From the definition of $P$, we can set:





\begin{equation}
 D^{i,j}_{B} = (PDJ+JDP+PIP)_{i,j} = \hat{D_{i}}+ \hat{D_{j}} - \hat{D}
\end{equation}

The row/columns mean of $D_{B}$ equal those of $D$ itself, and therefore $D_{B}$ solves $\|B\|_F^2$ with zero error. Since $JD_{B}J=0$, $D' = D_{A} + D_{B}$ is a minimizes both $\|A\|_F^2$ and $\|B\|_F^2$:

\begin{equation}\label{eq:lower_bound_def}
\begin{gathered}
  D_{A} + D_{B} = \argmin_{D'} \| D - D' \|_F^2\\ \textrm{s.t.} -JD'J \; \textrm{is PSD}
\end{gathered}
\end{equation}

With an error of:

\begin{equation}
  \mathbf{L} = \| D - (D_{A}+D_{B}) \|_F^2 = \sum_{i: \lambda_i < 0} \lambda_i^2
\end{equation}

Because equation \ref{eq:lower_bound_def} has only a single constraint out of the $3$ required from Euclidean matrices, $\mathbf{L}$ is a lower bound for the complete problem $\mathbfcal{L^{*}}\geq\mathbf{L}$. 

\subsection{Error Upper Bound}\label{subsection:upper_bound}

The truncation step at equation \ref{eq:truncation} changes the diagonal elements and $D_{A} + D_{B}$ (equation \ref{eq:lower_bound_def}) is strictly not hollow nor non-negative when $D$ is not Euclidean. However, we can consider what can be added to $D_{A}$ such the resulting matrix is hollow and non-negative but without incurring additional penalty on $\|A\|_F^2$. To this end, we will make use of the following lemma:

\begin{lemma}\label{unique_hollow}

For a given symmetric matrix $A$, there is unique symmetric matrix $B$ such that $A+B$ is hollow and $JBJ=0$.

\begin{proof}

From the constraint:

\begin{equation}
  JBJ = 0 \Longrightarrow J(BJ) = (JB)J= 0
\end{equation}

That is that $BJ$ is in the right null space of $J$ and $JB$ is in the left null space of $J$. The former implies that the rows of $BJ$ are constant and can be written as:

\begin{equation}\label{eq_left_null}
  BJ = B(I-P) = \textbf{1}_{N}\cdot c^{\top} \Longrightarrow B = BP + \textbf{1}_{N}\cdot c^{\top}
\end{equation}

Where $c$ is a columns vector. Multiplying equation \ref{eq_left_null} on the left by $P$ we get:

\begin{equation}
  PB = PBP + P\textbf{1}_{N}\cdot c^{\top} = \hat{B}+\textbf{1}_{N}\cdot c^{\top}
\end{equation}

Where $\hat{B}$ is a matrix filled with the mean value of $B$. Which in turn implies:

\begin{equation}
  c^{\top} = \frac{1}{N}\textbf{1}_{N}^{\top}B - \hat{B}\textbf{1}_{N}^{\top}
\end{equation}

So by the symmetry of $B$ we arrive at:

\begin{equation}
  B = \textbf{1}_{N}\frac{\textbf{1}^{\top}B}{N}+\frac{B\textbf{1}_{N}}{N}\textbf{1}_{N}^{\top}-\hat{B}
\end{equation}

$B\textbf{1}$ is a column vector, so to fit this constraint, $B$ must be in the form of:

\begin{equation}
  B = \textbf{1}_{N}\frac{b^{\top}}{N}+\frac{b}{N}\textbf{1}_{N}^{\top}-\hat{b}\frac{\textbf{1}_{N}\textbf{1}_{N}^{\top}}{N}
\end{equation}

Therefore, we have only $n$ degrees of freedom with which to find $B$. However, since $A+B$ has diagonal 0, this produces a series of $N$ equations which the diagonal elements of $B$ must fit:

\begin{equation}\label{eq:diag_solve}
  b_{i} - \frac{1}{2}\hat{b} = -\frac{1}{2}A_{i,i}
\end{equation}

The matrix describing this linear system has ones along its diagonals while its top triangular section is filled it $-\frac{1}{2\cdot N}$. Such a system is well-defined and therefore ascribes a unique solution onto $b$.

\end{proof}
  
\end{lemma}

According to lemma \ref{unique_hollow}, for the matrix $D_{A} = - \sum_{i}{\lambda_i\geq0}\lambda_iv_{i}v^{\top}_{i}$ there is a unique matrix $G=g\cdot1^{\top}+1\cdot g^{\top}-\hat{g}$ such that $D_{A}+G$ is hollow and $JGJ=0$. From the construction, we find that the diagonal elements of $D_{A}$ are:

\begin{equation}
  D^{i,i}_{A} = -\sum_{j:\lambda_{j}>0}\lambda_{j}\cdot v_{i,j}^2
\end{equation}

So by solving equation \ref{eq:diag_solve}:

\begin{equation}\label{g_eq}
\begin{gathered}
  g_{i} = \frac{1}{2}(\sum_{j: \lambda_{j}>0}\lambda_{j}\cdot v_{i,j}^2 + \frac{1}{N}\hat{g})\\
  \hat{g} = \frac{1}{n}\sum g_{i} = \frac{1}{N}\sum_{i}\sum_{j: \lambda_{j}>0}\lambda_{j}\cdot v_{i,j}^2
\end{gathered}
\end{equation}

We get that the matrix $G$ is:

\begin{equation}
  G_{i,j} = -\frac{D^{i,i}_{A}+D^{j,j}_{A}}{2} 
\end{equation}

While lemma \ref{unique_hollow} assists us with conforming to the hollowness constraints, we must also adhere to non-negativity. To this end, we will also apply the lemma shown below:

\begin{lemma}\label{non-negative}

For a negative-semi-definite (NSD) matrix $M$, the matrix $Q=M+C$ where:

\begin{equation}
  C_{i,j} = -\frac{M_{i,i} + M_{j,j}}{2}
\end{equation}

Has strictly non-negative values:

\begin{equation}
  Q_{i,j} \geq 0 
\end{equation}

\begin{proof}

The matrix $W=-M$ is PSD, and so is every $2\times2$ principle submatrix of $-M$ which implies \cite{horn2012matrix}:

\begin{equation}
  W_{i,j} \leq \sqrt{W_{i,i}W_{j,j}} 
\end{equation}

From the AM-GM inequality:

\begin{equation}
  W_{i,j} \leq \frac{W_{i,i}+W_{j,j}}{2}
\end{equation}

And since $ W_{i,j} = -M_{i,j}$ we arrive at:

\begin{equation}
  M_{i,j} \geq \frac{M_{i,i}+M_{j,j}}{2}
\end{equation}

And finally:

\begin{gather}
Q_{i,j} =M_{i,j}+C_{i,j} = M_{i,j} - \frac{M_{i,i}+M_{j,j}}{2} \geq 0 
\end{gather}

\end{proof}
  
\end{lemma}

From lemma \ref{non-negative}, the matrix $D_{A} + G$ fits all $3$ constraints, since it is also hollow, non-negative and $F=-J(D_{A} + G)J = -JD_{A}J$ is PSD.

While the resulting $G$ matrix does not change the value of $\|A\|_F^2$, it differs from $D_{B}$ and incurs an error in the $\|B\|_F^2$ term. The sum of the untruncated version of $JDJ$ and the matrix $PDJ + JDP + PDP$ equals to $D$ (equation \ref{eq:ID}) and is hollow. From lemma \ref{unique_hollow}, the latter matrix can be expressed similarly to equation \ref{g_eq} by including the negative eigenvalues:

\begin{equation}
\begin{gathered}
  \Tilde{g}_{i} = \frac{1}{2}(\sum_{j}\lambda_{j}\cdot v_{i,j}^2 + \frac{1}{N}\hat{\Tilde{g}})\\
  \Tilde{\hat{g}} = \frac{1}{N}\sum g_{i} = \frac{1}{N}\sum_{i}\sum_{j}\lambda_{j}\cdot v_{i,j}^2
\end{gathered}
\end{equation}

So if we define:

\begin{equation}
\Delta g_{i} = \frac{1}{2}\sum_{j: \lambda_{j}<0}\lambda_{j}\cdot v_{i,j}^2
\end{equation}

We get that the incurred error in the $\|B\|_F^2$ term is:

\begin{equation}
\|B\|_F^2 = \|\Tilde{G} - G\|_F^2 = \sum_{i,j}(\Delta g_{i}+\Delta g_{j})^2
\end{equation}

Which, included with the truncation error $\|A\|_F^2$, results in an upper bound to the complete problem:

\begin{equation}
  \begin{split}
  \mathbfcal{L^{*}}\leq\mathbf{U} = \sum_{i,j}(\Delta g_{i}+\Delta g_{j})^2 + \sum_{j: \lambda_{j}<0}\lambda_{j}^2
  \end{split}
\end{equation}

\section{Guaranteed Optimal Embeddings for non-Euclidean Distances}\label{subsection:ProjectedGradientDescent}

As formulated in equation \ref{eq:convex_problem}, the problem of finding the closest valid EDM to an arbitrary distance matrix involves minimizing a convex function subject to convex constraints. Formally, the constraint of $D'$ being an EDM is the intersection of three convex sets, and therefore, projected gradient descent \cite{boyd2004convex} (PGD) is guaranteed to converge to the global minimum $\mathbfcal{L^{*}}$:
\hfill \break
\begin{itemize}[noitemsep]
 \item ($C_{1}$) Matrices $D'$ such that $F=-JD'J$ is PSD 
 \item ($C_{2}$) Hollow Matrices $D'_{i,i} = 0 \; \forall i$
 \item ($C_{3}$) Non-Negative Matrices $D'_{i,j} \geq 0 \; \forall i,j$
\end{itemize}
\hfill \break
Starting from some initial value $D'_{0}$, each gradient descent step is defined as:

\begin{equation}\label{eq:projectedGD}
  D'_{t+1} = \text{proj}_{C_{1}\cup C_{2}\cup C_{3}}(D'_{t}-\varepsilon_{t}\nabla S(D'_{t}))
\end{equation}

Where $S$ is the error term in equation \ref{eq:convex_problem} and $\varepsilon_{t}$ is the step size at iteration $t$. In subsection \ref{subsection:lower_bound}, we have demonstrated how to project a matrix $K$ onto the set of matrices where $-JKJ$ is PSD:

\begin{equation}
  K_{C_{1}} = \text{proj}_{C_{1}}(K) = M + PK + (PK)^{\top} + PKP
\end{equation}

Where $M$ is the truncation of $-JKJ$ to its positive eigenspace:

\begin{equation}
  M = -\sum_{i: \lambda_{i}>0}\lambda_iv_{i}v^{\top}_{i}
\end{equation}

And $\{\lambda_{i},v_{i}\}_{i=1}^{N}$ are the eigenvectors and values of $F=-JKJ$. Projecting a matrix $K$ onto the set of hollow matrices is simply stetting the diagonal elements to $0$:

\begin{equation}
  K_{C_{2}} = \text{proj}_{C_{2}}(K) = K - \text{Diag}(K)
\end{equation}

And so is projecting onto non-negative matrices straightforward thresholding:

\begin{equation}
  K_{C_{3}} = \text{proj}_{C_{3}}(K) =   \begin{cases}
   K_{i,j} & \text{if $K_{i,j}\geq0$}\\
   0 & \text{otherwise}
  \end{cases}  
\end{equation}

To project onto the intersection of $C_{1}$, $C_{2}$ and $C_{3}$ we can use Dykstra's algorithm \cite{boyle1986method, gaffke1989cyclic}, which iterativly projects onto each convex set using corrective variables to find the element closest to $K$ in $C_{1} \cup C_{2} \cup C_{3}$. 

Using projected gradient descent (PGD) as described in equation \ref{eq:projectedGD} we are sure to find the closest valid EDM to an arbitrary distance matrix $D$ (Algorithm \ref{alg:Projected}). We can then apply classical MDS to produce embeddings whose distance match the idea euclidean matrix $D'$. The error of this solution is guaranteed to be within the lower and upper bounds derived in subsections \ref{subsection:lower_bound} and \ref{subsection:upper_bound}.

Despite its theoretical soundness, there are hurdles preventing this algorithm from being applied on real-world datasets. It relies on having access to the complete distance matrix $D$, which not trivial for large datasets and complex metrics such as the Wasserstein Distance. This algorithm is indeed infeasible on large cohorts, which motivates our stochastic Wormhole approach. 

\subsection{Theoretical Results}\label{subsection:theory_results}

On small examples, we can however implement algorithm \ref{alg:Projected} and empirically show that its solution is within our derived error bounds. We construct an a cohort of empirical distribution whose pairwise Wasserstein distance is non-Euclidean based on the example presented in section 8.3 in \cite{peyre2019computational}. We start by generating a set of distributions in $R^{2}$ with support on $x_{1} = (0,0), x_{2} = (1,0), x_{3} = (1,0), x_{4} = (1,1)$ where the probability of each points is either $\{0,\frac{1}{4},\frac{2}{4},\frac{3}{4},1\}$ such that the weights sum to $1$ for each distribution, which produces $N=35$ weighted point clouds.

\begin{wrapfigure}{l}{0.55\textwidth}
\includegraphics[width=0.55\textwidth]{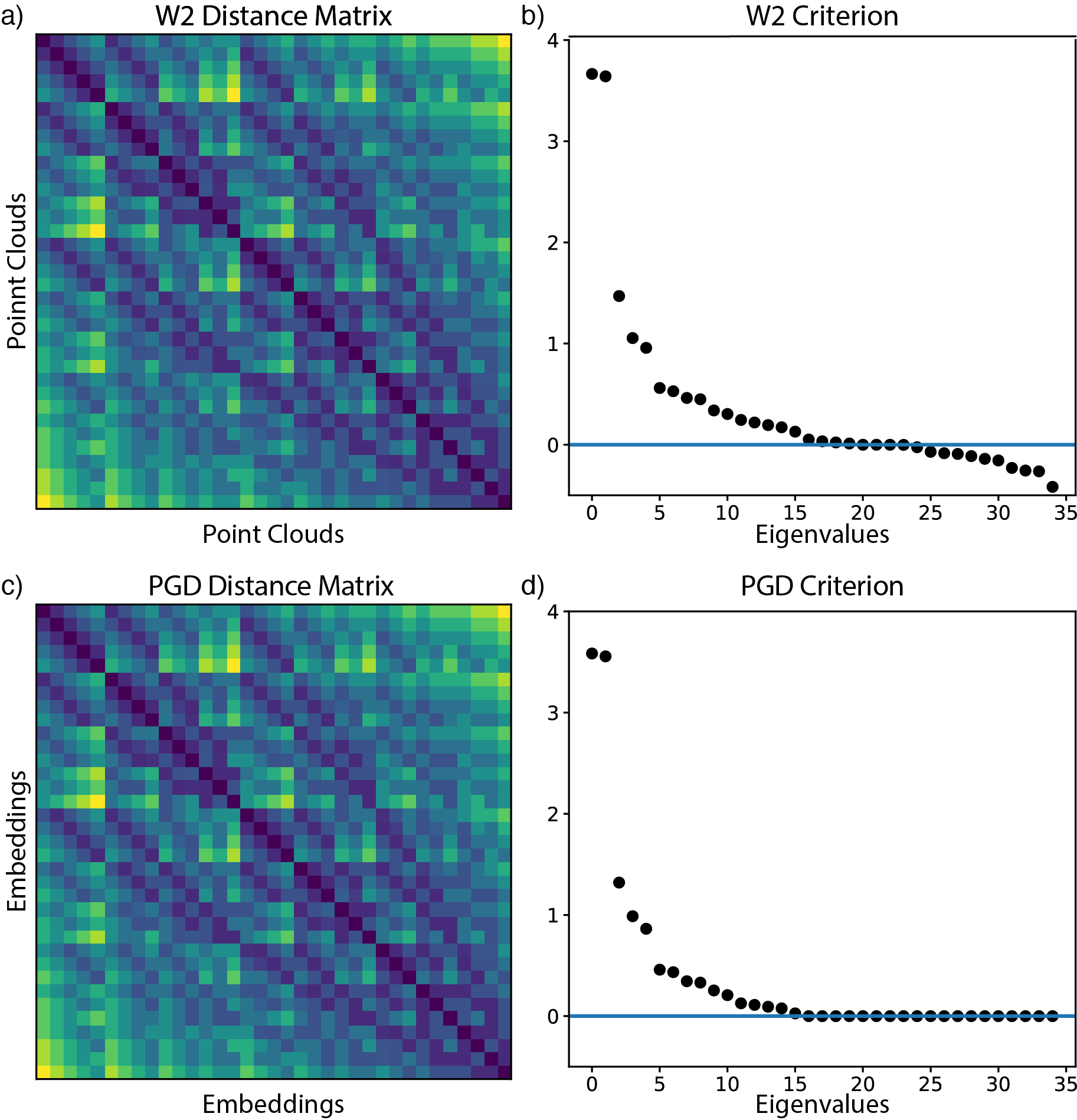}
\caption{\textbf{Non-Euclidean Wasserstein distance matrix and PGD convergence to it. a.} Pairwise Wasserstein distances between the 35 \textit{Simplex} point clouds. \textbf{b.} Spectrum of the double centered Wasserstein distance matrix (i.e. $F=-JDJ$). Since there are negative eigenvalues, the original distance matrix is non-Euclidean. \textbf{c.} Distance matrix produced PGD algorithm \ref{alg:Projected} to fit a valid EDM. \textbf{d.} Spectrum of the double centered PGD distance matrix. The PGD distance matrix is an EDM since it is hollow, positive and its criterion matrix is PSD.}
\label{suppfig:TheoryPGDFigure}
\vspace{-2em}
\end{wrapfigure}

To fit within our framework of empirical distributions, we convert each weighted point cloud to a uniform one by duplicating each point based on its probability. For example, if a point is weighted $\frac{3}{4}$, we duplicate it so that it appears thrice with uniform weight. We then add a small amount of independent Gaussian noise to make the points in each cloud are all different from one another. This scheme produces a set of 35 point clouds each with 4 points whose pairwise Sinkhorn divergence distance matrix is non-Euclidean (Figure \ref{suppfig:TheoryPGDFigure}a-b). Since original point weights are derived from the 4 
dimensional simplex, we dub this the \textit{Simplex} example.

We apply Algorithm \ref{alg:Projected} on the \textit{Simplex} distance matrix and find that the resulting error fits well within our derived bounds (Table \ref{table:bounds} and Figure \ref{suppfig:TheoryPGDFigure}c-d). We similarly apply our Wormhole algorithm to produce embeddings and classical MDS with embedding dimension of $35$ (Table \ref{table:bounds}). Despite the stochastic training of Wormhole, its error is close to the global minima derived by PGD. As expected, cMDS performs poorly and its error is outside of the upper bound.

We compare Wormhole to PGD and cMDS on two more non-Euclidean examples. The first, \textit{Gaussians}, we generate by randomly producing $N=128$ mean vectors ($\mu$) of length $d=8$ and $N$ covariance matrices ($\Sigma$) of shape $d\times d$. Both $\mu$ vectors and $\Sigma$ matrices are sampled from the standard normal, where $\Sigma$ matrices are ensured to valid covariances (i.e. PSD) by multiplying resulting matrices with their transpose. From each of the $N$ resulting parameterized Gaussians we randomly samples 512 points to create point clouds. Values are then normalized to be between $-3$ and $3$. The pairwise Sinkhorn divergence between these point clouds is not Euclidean as the minimum eigenvalue of the resulting criterion matrix was $\lambda_{min}=-0.051$.

\begin{wrapfigure}{r}{0.4\textwidth}
\includegraphics[width=0.4\textwidth]{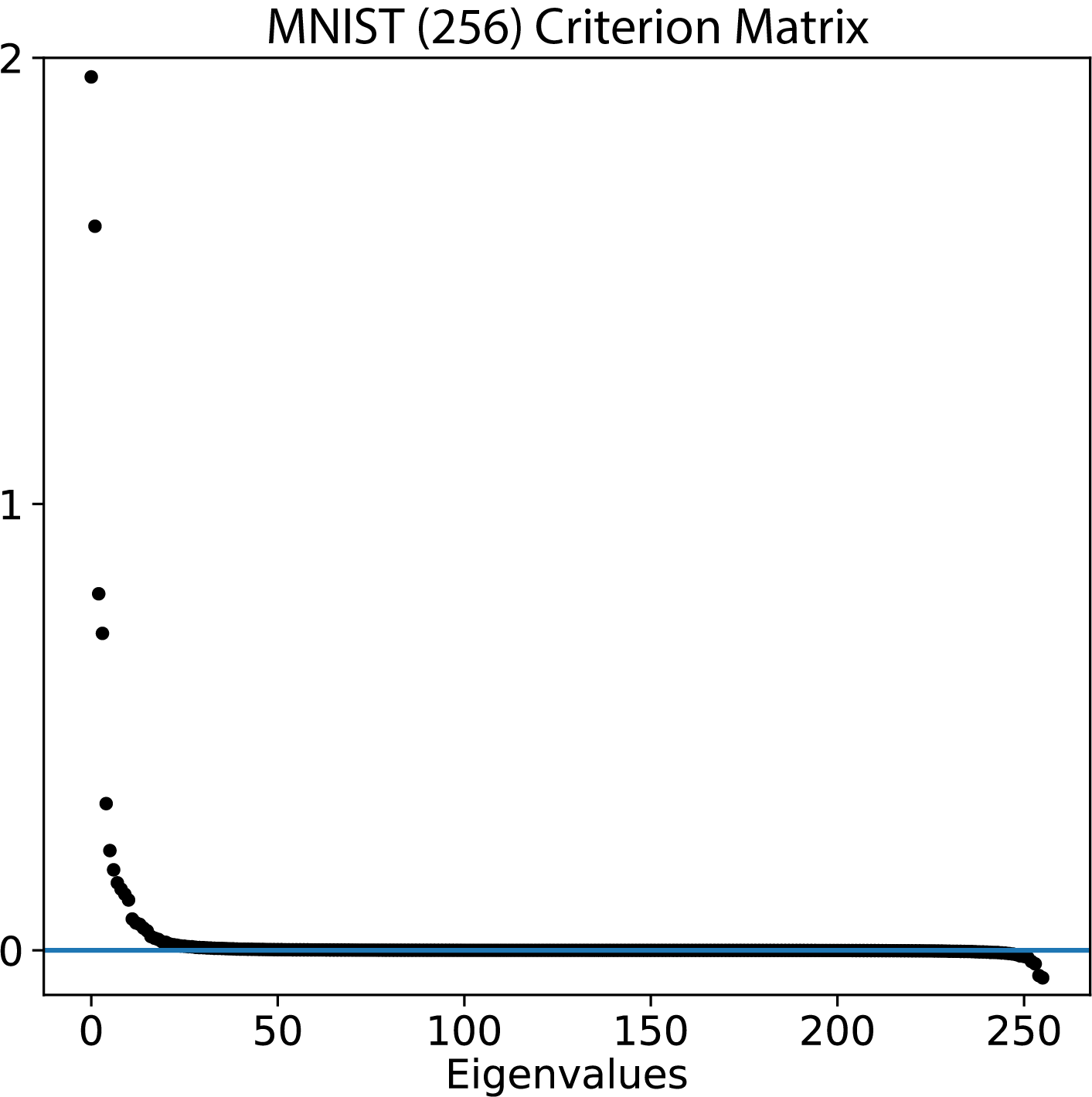}

\caption{\textbf{Spectrum of the criterion matrix from real data}. On the randomly sampled \textit{MNIST}, we calculated the $256\times 256$ pairwise Wasserstein distance matrix and the eigenvalues of its criterion matrix. While $D$ is not Euclidean, as $F=-JDJ$ has negative eigenvalues, their contribution to the spectrum is not as prominent relative to the \textit{Simplex} case (Figure \ref{suppfig:TheoryPGDFigure}).}
\label{suppfig:mnist_spectrum}
\vspace{-3em}
\end{wrapfigure}

An important insight from our upper and lower bounds derived in subsections \ref{subsection:lower_bound} and \ref{subsection:upper_bound} is that our Wormhole algorithm is expect to perform well in cases where the negative eigenspace is considerably smaller than its positive counterpart. To 
demonstration of the non-Euclidean nature of the Wasserstein distance in our real-world datasets and explain the accuracy of our approximation nevertheless, we draw $N=256$ point clouds from the binarized \textit{MNIST} dataset. The pairwise Sinkhorn divergence matrix was non-Euclidean, and its criterion matrix did have negative eigenvalues $\lambda_{min}=-0.062$. However, unlike the \textit{Simplex} example, on this sample of real dataset the negative eigenspace was much smaller in magnitude and less prominent (Figure \ref{suppfig:mnist_spectrum}).

On both the \textit{Gaussian} and \textit{MNIST} examples, we find that both Wormhole and PGD error terms are within the lower and upper error bounds (Table \ref{table:bounds}). We again stress that the PGD algorithm \ref{alg:Projected} is not realizable on large datasets, and the stochastic nature of Wormhole enables its scalability, albeit at a marginal cost compare to optimal solution.

\section{Implementation Details}\label{section:imp_details}

We made use of three different Wasserstein approximation algorithms in our paper, DiffusionEMD, Fréchet distances and our novel Wormhole. In the following, we describe how each approach was applied.

\subsection{Wasserstein Wormhole}

\subsubsection{Pre-processing}

Stability and efficiency of Wasserstein distance computations, specifically entropically regularized OT, is highly dependent on the scale of the input. The larger the values in the ground-cost distance matrix, the more iterations of Sinkhorn are required and the likelihood of machine-point precision error (\textit{nan}) grows. 

To mitigate \textit{nan}-related risks and to ensure swift training of Wormhole, by default we scale the point cloud values to reside between $-1$ and $1$ via min-max normalization. For high-dimensional distributions, such as the \textit{Cellular Niche} cohort in subsection \ref{subsection:niches}, we further scale down by a factor of square-root number of dimensions. To accommodate instances where no re-scaling is desired, this operation can be easily disabled by the user.

\subsubsection{Neural Architecture}

Wasserstein Wormhole uses Transformers to both encode point clouds into embeddings, and reconstruct them from the embedding using the decoder. By default, the encoder consists of an \textit{embedding} layer to transform the coordinate into the dimension of the encoding. This is followed by $3$ attention blocks which consist of $4$-headed attention followed by a $2$-layer fully connected (FC) network. The embedding dimension is set to be $128$ and the size of the hidden layer in FC network within the attention block is $512$ neurons. To produce the permutation-invariant embedding, we average the attention encodings of each sample within the point cloud.

The decoder only has purview onto the embedding, and thus lacks explicit knowledge of the number of samples within the original empirical distribution. Since the decoder is trained via Wasserstein distance with respect to the input point cloud, there is not constraint on decoder to produce an output with the same number of points. For simplicity, we set the decoder to produce a empirical distribution with $m$ points, where $m$ is the median sample size of cohort, as noted in Table \ref{table:summary}.

Given an encoding, the decoder first applies a linear, \textit{multiplier} layer, which returns $m$ new embeddings, one for each point in the output. From there, the default decoder architecture mirrors the encoder, with $3$ attention blocks followed by an \textit{unembedding} layer to project back into the original point cloud dimension. 

\subsubsection{Training and OT Parameters}

Wormhole is trained with $10,000$ gradient descent (GD) steps using the ADAM optimizer \cite{kingma2014adam} with an initial learning rate of $10^{-4}$ and an exponential decay schedule. The default batch size was set to $16$, which might seem small, but requires $\binom{16}{2}=120$ computations of Wasserstein distance in each iteration step to compute the encoding loss. With the default settings, as long as the cohort consists of more than $1600$ point clouds, Wormhole requires fewer OT computations to derive embeddings compared to direct pairwise Wasserstein.

Convergence speed of Sinkhorn iteration depends to the level of entropic regularization, with larger values of $\varepsilon$ (equation \ref{eq:w2}) resulting in quicker convergence. Within Wormhole, we have two choices for $\varepsilon$, one for the encoder and another for the decoder. Since the former loss requires many more computations, its default is $\varepsilon_{enc}=0.1$, while we set the regularization at the decoder to a smaller value of $\varepsilon_{dec}=0.01$.

In all experiments of Wormhole, we used the Sinkhorn divergence loss (equation \ref{eq:s2}) instead of standard entropic Wasserstein (equation \ref{eq:w2}). 
This includes the GW examples, where we used its Sinkhorn divergence equivalent. These choice reflects our desire that similar point-clouds have close embeddings, rather than their distances reflecting self-transport cost. We additionally implemented other OT distances which users of Wormhole can choose from. Namely, we included the standard (entropic) Wasserstein and GW distances, and also EMD ($W_{1}$) and its Sinkhorn divergence $S_{1}$, which use $L_{1}$ ground-distance instead of $L_{2}$.

To calculate the object class accuracy of learned Wormhole embeddings, we train an MLP classifier using the default implementation from scikit-learn \cite{pedregosa2011scikit}, on the train-set encodings and labels. We opt to a neural network instead of other alternatives such as kNN or Random-Forest due to the dimensionality of the embeddings. We report accuracy of the classifier's predictions the test-set encodings and their true labels.

\subsection{DiffusionEMD}

The key concept behind DiffusionEMD is the Wavelet approximation to EMD. Briefly, \cite{shirdhonkar2008approximate} showed that the Euclidean distance between Wavelet \cite{torrence1998practical} decomposition of an 2D voxelized histograms (i.e. gray-scale image) is correlated with their $W_{1}$ distance. By applying the spectral analogous of Wavelets, DiffusionEMD extends \cite{shirdhonkar2008approximate} from images to graphs, and generalize it to more than $2$ dimensions and requiring fewer constrains on the ground space.

Applying DiffusionEMD requires first constructing a graph of the support of the distribution. For images, such as MNIST and FashionMNIST, this meant building a graph with $28\times28=784$ nodes, one for each pixel, and edges are connected nodes if distance between their pixels is smaller than $\sqrt{2}$. For 3D point clouds from ModelNet40 and ShapeNet, we first divided their spaces into $16^{3}=256$ voxels, and nodes were connected in the graph if they are closer than $\sqrt{3}$ units away.

\begin{wraptable}{l}{0.5\textwidth}
\begin{tabular}{l c c}
\hline
Dataset  & Wormhole & DWE \\ [0.5ex] 
\hline\hline
MNIST & $\mathbf{2.57\cdot10^{-5}}$ & $1.63\cdot10^{-4}$  \\
FashionMNIST & $\mathbf{1.41\cdot10^{-5}}$ &$1.88\cdot10^{-4}$  \\
ModelNet40 & $\mathbf{8.89\cdot10^{-5}}$ & $3.52\cdot10^{-4}$ \\
ShapeNet & $\mathbf{1.14\cdot10^{-4}}$ & $6.66\cdot10^{-4}$ \\
\hline\hline
Rotated ShapeNet (GW) & $\mathbf{2.28\cdot10^{-3}}$ & NA \\
\hline\hline
MERFISH Cell Niches & $\mathbf{6.94\cdot10^{-3}}$ & OOM \\
SeqFISH Cell Niches & $\mathbf{9.90\cdot10^{-4}}$ & OOM \\
scRNA-seq Atlas  & $\mathbf{3.80\cdot10^{-5}}$ & OOM\\
\hline
\end{tabular}
\caption{Test-set Mean Squared Error (MSE) measuring the discrepancy between embedding distances and  Wasserstein distances. Values are derived by $10$ random instances of sampling $128$ point clouds for each dataset and MSE is computed between the embedding and OT pairwise distance matrices. Since DiffusionEMD only produces embedding which are only correlated OT, it was not included in this benchmarking. Similarly, Fréchet is designed to approximate the unregularized OT, while Wormhole and DWE both are trained on entropic OT, so Fréchet was excluded as well. NA, Not Applicable. OOM, Out-Of-Memory.}
\vspace{-3em}
\label{table:MSE}
\end{wraptable}

Each point cloud is then considered a distribution over the graph, which is based on the number of points within each voxel or grid point. We applied DiffusionEMD with default parameters following instruction in \url{github.com/KrishnaswamyLab/DiffusionEMD}. DiffusionEMD calculates the spectral wavelet decomposition which acts as its OT preserving embeddings. Object classification is based on a kNN ($k=1$) classifier between train and test-set embeddings, repeating the process from the original manuscript.

This approach can be applied for general point clouds, and does not necessarily require voxelization. However, since every single point among all point clouds would then require its own node in the graph, DiffusionEMD will not scale in that case. This prevented us from applying it on the two high-dimensional \textit{Cell Niches} and scRNA-seq datasets, and DiffusionEMD produces an out-of-memory (OOM) errors.

\subsection{DWE}\label{subsection:DWE}

Deep Wasserstein Embeddings (DWE) \cite{courty2017learning} is a CNN based embedding algorithm for gray-scale images. Conceptually, DWE uses a Siamese-training scheme, where pairs of images are encoded simultaneously by a convolutional auto-encoder, and the network is trained so that distances between embeddings match the Wasserstein Distances between the images. A decoder is also trained to minimize the KL divergence between output and input images.

This generates a parametric embedding scheme that reduces the computational complexity for Wasserstein based analysis. However, the CNN backbone limits the applicability of DWE, and it can only embed 2D images, or point-clouds whose support is a fixed-grid. Indeed, DWE cannot embed cohorts of point-clouds in arbitrary space and high-dimensions which cannot be voxelized.

The original code-base for DWE is written in \textit{python2}, and is unrunnable on modern machines. To efficiently benchmark DWE against Wormhole and other Wasserstein acceleration approaches, we re-wrote the DWE code in \textit{python3} and JAX \url{github.com/DoronHav/jax\_dwe}. We also extended DWE from 2D images to 3D grid by adding 3D convolutional layers. MNIST and FashionMNIST, already being 2D gray-scale images, did not require any processing for DWE to be applied. Being 3D point-clouds, the ShapeNet and ModelNet40 cohorts were first voxelized into a $16^{3}=256$ grid. All other datasets were too high-dimensional for DWE and voxelization produced OOM errors. 

The optimization of DWE followed the same procedure as Wormhole, where the model was trained for 10,000 gradient descent steps with the ADAM optimizer and an exponential decay schedule. To match the per-step Wasserstein comparisons of Wormhole, the batch size for DWE was $128$. Like Wormhole,  DWE embeddings were trained to reconstruct Sinkhorn divergence to allow for direct comparisons between the algorithms. Despite similar training parameters, Wormhole recovered OT space more reliably than DWE on low-dimensional datasets (Tables \ref{table:summary}, \ref{table:MSE}).

\subsection{Fréchet}\label{subsection:frechet_details}

Acting as a naive baseline, this approach utilizes the closed form solution OT has in the Gaussian case as an approximation for general point clouds. For all empirical distribution in a cohort, we calculate their $d$-dimensional mean and $d\times d$ covariance matrix. We then compute the Fréchet distance (equation \ref{eq:frechet}) between each pair to estimate their Wasserstein distance.

The cost of each Fréchet computation is $O(d^{3})$. As long as $d\ll m$, where $m$ is the number of points in each cloud, Fréchet is significantly less computationally intensive than Sinkhorn iterations. On high dimensional datasets, $d\centernot\ll m$, and Fréchet can not be applied.

This approach does not provide us with an embedding. To still 
assign test-set with label predictions, we manually perform a kNN ($k=30$) classifier based on Fréchet distances. We calculate the approximated distance for each test-set point cloud to every train-set example and assign it the most common label among its $k$ closet neighbours.

\subsubsection{Gaussian Gromov-Wasserstein}

For GW, the formula written in equation \ref{eq:frechet} no longer applies. While a closed form solution for GW in the Gaussian case has not been found, \cite{delon2022gromov} derived a lower bound to it, which we used in lieu. For two Gaussians $N(\mu_{1},\Sigma_{1})$ and $N(\mu_{2},\Sigma_{2})$ in $d$ dimensions, their GW distance is greater or equal to:

\begin{gather}\label{eq:normal_GW}
\begin{split}
  LGW^{2} &= 4(\text{Tr}(D_{1})-\text{Tr}(D_{2}))^{2}
      +4(\|D_{1}\|_F-\|D_{2}\|_F)^{2}\\ 
      &+4(\|D_{1}\|^{2}_F-\|D_{2}\|^{2}_F)
      +4(\|D_{1}-D_{2}\|^{2}_F)\\ 
\end{split}
\end{gather}


Where $D_{i}$ is the diagonalization of $\Sigma_{i}$, sorted in decreasing order. Using equation \ref{eq:normal_GW}, we estimate pairwise GW distances between Rotated ShapeNet point cloud, which we assign test-set labels using the same kNN classification scheme described in subsection \ref{subsection:frechet_details}


\end{document}